\documentclass{article}

\usepackage{algorithm}
\usepackage{algpseudocode}
\usepackage{enumitem}

\usepackage[preprint]{neurips_2025}

\usepackage[utf8]{inputenc} 
\usepackage[T1]{fontenc}    
\usepackage{hyperref}       
\usepackage{url}            
\usepackage{booktabs}       
\usepackage{amsfonts}       
\usepackage{nicefrac}       
\usepackage{microtype}      
\usepackage{xcolor}         
\usepackage{graphicx}
\usepackage{amsmath}
\usepackage{makecell}

\title{\Large Towards Universal Semantics with Large Language Models}

\author{
    Raymond Baartmans\textsuperscript{1}, Matthew Raffel\textsuperscript{1}, Rahul Vikram\textsuperscript{1}, Aiden Deringer\textsuperscript{2}, Lizhong Chen\textsuperscript{1} \\ 
    \textsuperscript{1}Oregon State University, \textsuperscript{2}Pennsylvania State University \\
    \texttt{\{baartmar, raffelm, vikramr, chenliz\}@oregonstate.edu} \\
    \texttt{\{abd5984\}@psu.edu} \\
}

\begin{document}

\maketitle

\begin{abstract}
The Natural Semantic Metalanguage (NSM) is a linguistic theory based on a universal set of \textit{semantic primes}: simple, primitive word-meanings that have been shown to exist in most, if not all, languages of the world. According to this framework, any word, regardless of complexity, can be paraphrased using these primes, revealing a clear and universally translatable meaning. These paraphrases, known as explications, can offer valuable applications for many natural language processing (NLP) tasks, but producing them has traditionally been a slow, manual process. In this work, we present the first study of using large language models (LLMs) to generate NSM explications. We introduce automatic evaluation methods, a tailored dataset for training and evaluation, and fine-tuned models for this task. Our 1B and 8B models outperform GPT-4o in producing accurate, cross-translatable explications, marking a significant step toward universal semantic representation with LLMs and opening up new possibilities for applications in semantic analysis, translation, and beyond. Our code is available at \hyperlink{https://github.com/OSU-STARLAB/DeepNSM}{https://github.com/OSU-STARLAB/DeepNSM}.
\end{abstract}

\section{Introduction} 
\label{sec:introduction}
Semantics, the study of meaning, lies at the center of human language and is vital for nearly all language-based tasks. However, representing meaning in a way that is precise, unambiguous, and transferable across languages remains a major challenge, since human languages are known to contain unique words and concepts that do not directly translate into others. Even seemingly universal words like "green" and "blue," have been shown to simply not exist in some languages \cite{wierzbicka_why_2008}. This problem is often compounded by conventional semantic approaches, such as dictionary definitions, which often suffer from circularity or rely on culturally specific terms that may seem intuitive to English speakers but are not universally applicable.

Modern large language models (LLMs) are overwhelmingly pretrained on English text, enabling strong general natural language processing (NLP) performance \cite{achiam2023gpt, touvron2023llama}. However, the dominance of modern English in LLM training data introduces challenges for tasks in low-resource domains that require precise semantic understanding, such as low-resource translation \cite{song2025llm} or legal text interpretation \cite{litaina2024towards, siino2025exploring}, as models may struggle to generalize beyond the language-specific semantic representations learned in their training. In low-resource translation, for example, approaches to address this semantic gap have tried leveraging multilingual data to learn more "universal" semantic representations during training \cite{man_lusifer_2025}. Other methods focus on test-time behavior, such as dynamically accessing external lexical resources, like dictionaries, to enable fine-grained semantic interpretation \cite{dimakis2024dictionary, merx2024low, elsner2023translating}. A lexical semantic framework that could serve as a basis for representing and reasoning about word meaning in a universal, unambiguous, and precise way would offer significant benefits for LLMs for not just low-resource translation, but a wide range of semantically demanding tasks.

\begin{figure}
    \centering
    \includegraphics[width=.9\linewidth]{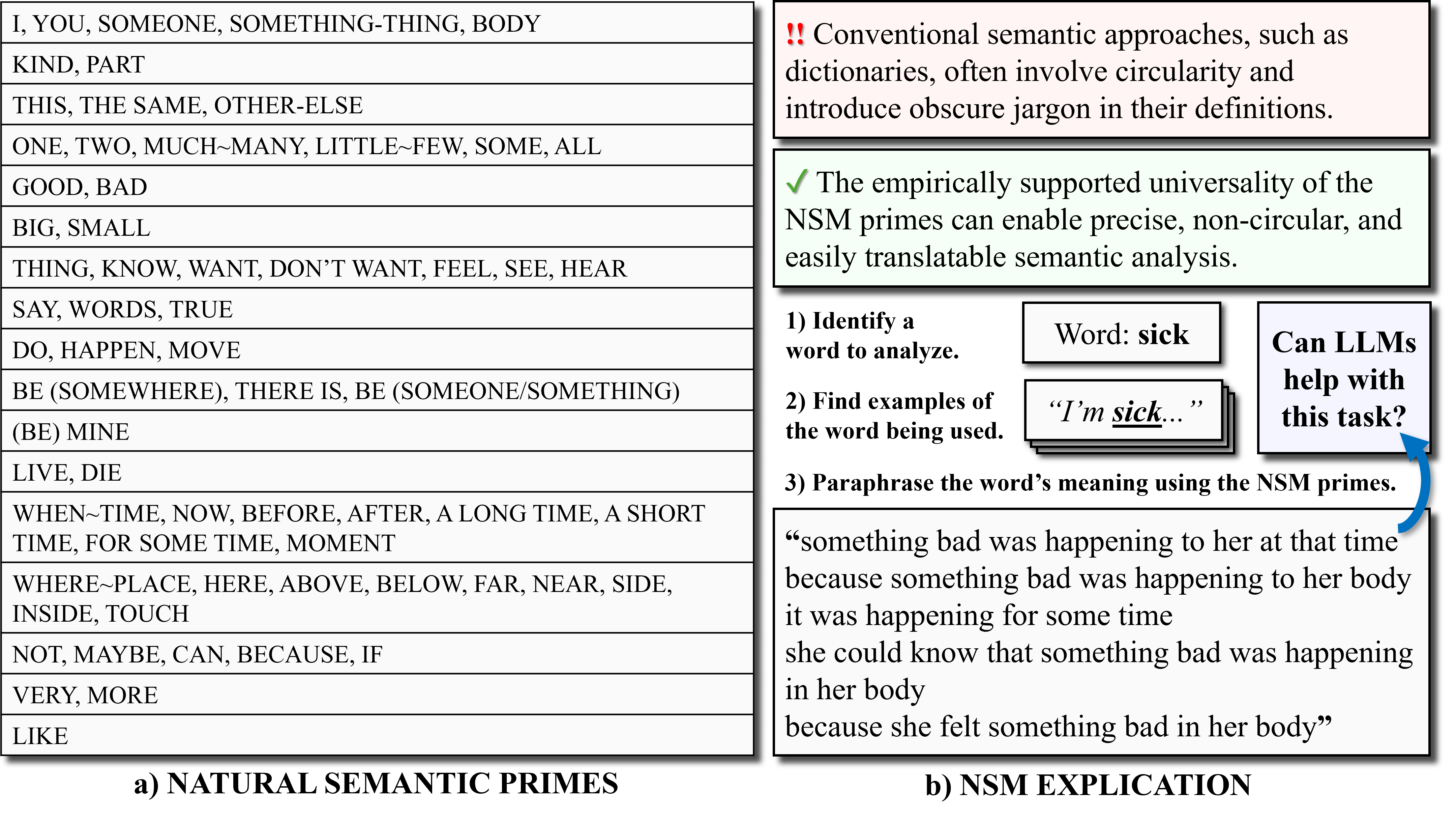}
    \label{fig:nsm_primes}
    \caption{A list of the proposed Natural Semantic Primes, along with an example demonstrating how the NSM framework is applied to paraphrase a word into an explication of its meaning. The pictured explication is adapted from \cite{goddard_semantics_2021}.}
\end{figure}

The Natural Semantic Metalanguage (NSM) \cite{goddard_1_2008} is a linguistic framework that proposes a small set of primitive, universal word-meanings, known as \textit{semantic primes} (Figure \ref{fig:nsm_primes}a). These primes are considered primitive because they represent fundamental semantic units that cannot be defined in simpler terms; their meanings are taken to be self-evident. Substantial empirical evidence shows that equivalent primes exist and are lexicalized (i.e., represented by specific words) in most, if not all, languages \cite{goddard_semantic_2010, nsm_approach, goddard_semantic_2010, goddard_semantic_2014, wierzbicka2021semantic}, enabling translation into any language without loss of meaning. Building on this, words, regardless of their complexity or cultural nuance, can be paraphrased using semantic primes (Figure \ref{fig:nsm_primes}b) to reveal a universally translatable meaning. These paraphrases, called explications, provide a clear, universal way to describe and compare word meanings without the circularity, jargon, or culture-specific assumptions of conventional semantic approaches like dictionary definitions. These advantages have led to the application of the NSM framework in many fields, including cross-cultural communication \cite{goddard_cross-linguistic_2008, goddard_semantic_2014}, cultural analysis \cite{wierzbicka_understanding_1997, bromhead_reign_2009, wierzbicka_experience_2010, goddard_cultural_2004}, language teaching \cite{arnawa2017implementation, sadow2022pedagogical}, and beyond \cite{goddard_evolutionary_2014, bromhead_applied_2023, zeifert2023natural}.

If scaled, the NSM framework could offer a foundation for universal semantic representation in LLMs and NLP systems, supporting a range of downstream tasks. However, drafting explications has traditionally been a slow, manual process, with experts sometimes taking weeks or months to produce a single explication \cite{goddard2009natural}. Surprisingly, no previous research has explored how machine learning can be used to automate or accelerate this process. In this work, we present the first investigation into using LLMs for universal semantic analysis within the NSM framework, focusing specifically on generating and evaluating NSM explications. We introduce the NSM framework and define a task setup (Section \ref{sec:nsm_for_semantic_analysis}), and identify three key challenges in adapting LLMs: the lack of effective models for generating NSM explications, the absence of a suitable dataset or benchmark, and the need for automated methods to evaluate the legality, quality, and cross-translatability of explications. Our contributions are as follows: 
\begin{enumerate}[itemsep=1pt, topsep=1pt]
    \item We propose the first automated methods for evaluating the legality, descriptive accuracy, and cross-translatability of NSM explications (Section \ref{sec:evaluation_methods}).
    \item We construct the first dataset for training and evaluating LLMs on this task, using our evaluation methods as a quality filter (Section \ref{sec:dataset}).
    \item We introduce DeepNSM, two fine-tuned LLMs (1B and 8B), and show that they outperform general LLMs such as GPT-4o and Gemini in explication quality (Section \ref{sec:model_experiments}).
    \item We show that NSM explications have significantly higher cross-translatability into low-resource languages than traditional semantic approaches, such as dictionary definitions, aligned with their proposed universality (Section \ref{sec:model_experiments}).
\end{enumerate}

The methods, datasets, and models introduced in this work constitute a pivotal first step in adapting LLMs for the NSM framework, establishing a foundation for future research. It opens the door to a universal semantic layer for AI, with far-reaching implications for linguistic study and real-world language applications. All code, models, and datasets are open-sourced \footnote{https://github.com/OSU-STARLAB/DeepNSM}.

\section{Natural Semantic Metalanguage} 
\label{sec:nsm_for_semantic_analysis}

\subsection{Overview of the NSM Approach}
\label{sec:nsm_overview}
Figure \ref{fig:nsm_primes}a presents the full list of English exponents of the NSM semantic primes, which have been empirically attested in over 90 languages and counting \cite{goddard_semantic_2010, nsm_approach, goddard_semantic_2010, goddard_semantic_2014, wierzbicka2021semantic}, including many low-resource and endangered languages. The NSM approach is based on the principle that the meaning of any word, regardless of its complexity, can be fully paraphrased using only the semantic primes. This approach can be applied to words,  multi-word expressions (MWEs), proverbs \cite{goddard_words_2014}, and longer texts \cite{wilson2023did}.

As illustrated in Figure \ref{fig:nsm_primes}b, the analytical process typically begins by selecting a target word and gathering contextual examples in which the word is used in the same way. Researchers then examine these examples to draft an explication, paraphrasing the word’s meaning using only NSM primes. This process is also known as reductive paraphrasing \cite{goddard_semantic_1998, goddard2009natural}, and follows a few core principles. The main guideline is to rely as much as possible on the semantic primes. However, explications, especially in early drafts, do not need to be fully reduced to primes to be useful. Explications are allowed to include non-prime words, referred to as \textit{semantic molecules}, to manage complexity. While some molecules are considered near-universal, many are culture-specific \cite{goddard_1_2008, bromhead_semantics_2020, maskova_semantic_2022-1, otomo_nsm_2005, wierzbicka_understanding_1997}. In all cases, the goal is to be as reductive and non-circular as possible. Molecules should only be used when truly necessary, for instance, to simplify a very long explication, should be simpler than the word being paraphrased, and ideally are words that have already been paraphrased in the semantic primes.

Once an explication is drafted, it is tested by substituting it for the target word in the original usage examples to check whether it fully preserves the target word’s meaning, captures its key entailments, and remains as minimal as possible (a.k.a. \textit{minimality}). In addition, the explication should be easy to translate across different languages and cultural contexts. Even if written in English, a paraphrase composed largely of semantic primes is less likely to be distorted in translation, as using the primes simplifies the translation task from a semantic challenge to primarily a grammatical and syntactical one.

\subsection{Applications of NSM}
\label{sec:nsm_applications}
Once a word has been paraphrased using semantic primes, the resulting explication can support a range of downstream applications. A key example is translation, particularly for low-resource languages without direct equivalents for many English terms.
When translating between two languages, a natural and principled first step is to identify what the two languages have in common. The NSM framework provides a theoretical foundation for this. 
For instance, translating "color" into Warlpiri \cite{wierzbicka_why_2008}, which lacks dedicated color terms, becomes easier when "color" is first reduced to a semantically transparent NSM explication, shifting the challenge from semantics to syntax and grammar. Similarly, the universality of semantic primes has led to the application of the NSM framework for language learning and language revival efforts \cite{arnawa2017implementation, sadow2022pedagogical}. Explications have also been used in literary analysis, such as examining historical texts for semantic drift and cultural changes \cite{wierzbicka_experience_2010, bromhead_reign_2009}. Automating explication generation could extend this application into traditional NLP tasks like sentiment analysis, document retrieval, or legal text analysis \cite{siino2025exploring, litaina2024towards}, where capturing subtle emotional or contextual meaning is crucial—an area where NSM representations could offer more transparent and precise semantic grounding.

The universality and interpretability of the semantic primes could also provide interesting applications for LLMs themselves. Models that generate text using semantic primes could make the outputs of LLMs and other AI models accessible to speakers of all languages. Explications could support more transparent and interpretable model reasoning by grounding outputs in culturally neutral terms. The NSM framework could even inform LLM architecture design, for example, using semantic primes as a foundation for human-interpretable embedding dimensions, as proposed in recent work \cite{rodgers_creating_nodate}.

\begin{figure}
    \centering
    \includegraphics[width=.9\linewidth]{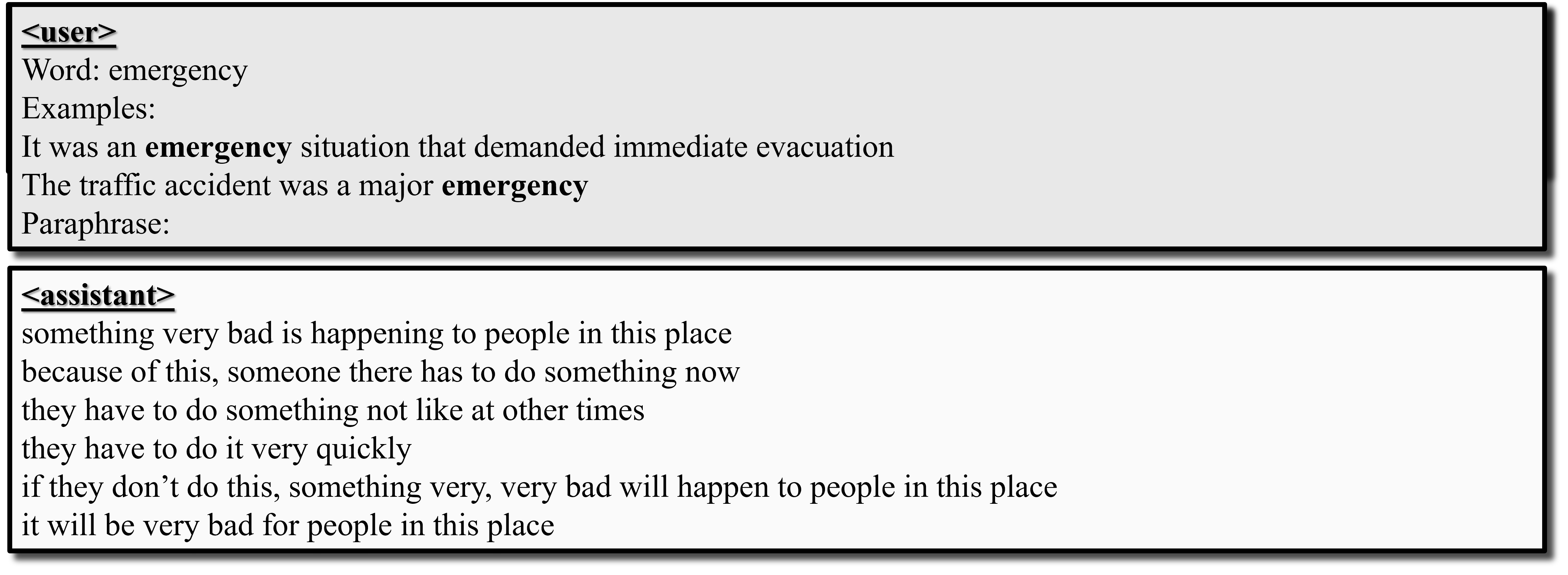}
    \caption{Envisioned interaction where a user prompts an LLM to generate an NSM-style explication of a target word given usage examples, as described in Section \ref{sec:task_setup}.}
    \label{fig:user_assistant_interaction}
\end{figure}

\subsection{Opportunity of LLM for NSM \& Task Setup}
\label{sec:task_setup}
While the NSM is an attractive and promising approach to assist NLP/ML tasks in numerous ways, the substantial time required to draft explications has been a critical limitation.
Because the process of reductive paraphrasing is not governed by strict formal rules, it resists automation through rule-based systems. As a result, even experienced NSM practitioners may spend weeks or months crafting a single explication \cite{goddard2009natural}. The potential for LLMs to assist in this process presents an opportunity to scale the NSM approach, address one of its key drawbacks, and integrate it into broader NLP systems, such as machine translation pipelines or language learning tools. We define a basic task setup for using LLMs to assist in generating NSM explications, shown in Figure \ref{fig:user_assistant_interaction}. In our formulation, a user selects a target word and provides several contextual examples illustrating its use. The goal is for the model to paraphrase the word’s meaning using only the semantic primes. In the following section, we examine the challenges of using LLMs to perform this task.

\subsection{Current Challenges in using LLMs for the NSM Approach}
\label{sec:challenges}

\begin{figure}
    \centering
    \includegraphics[width=0.9\linewidth]{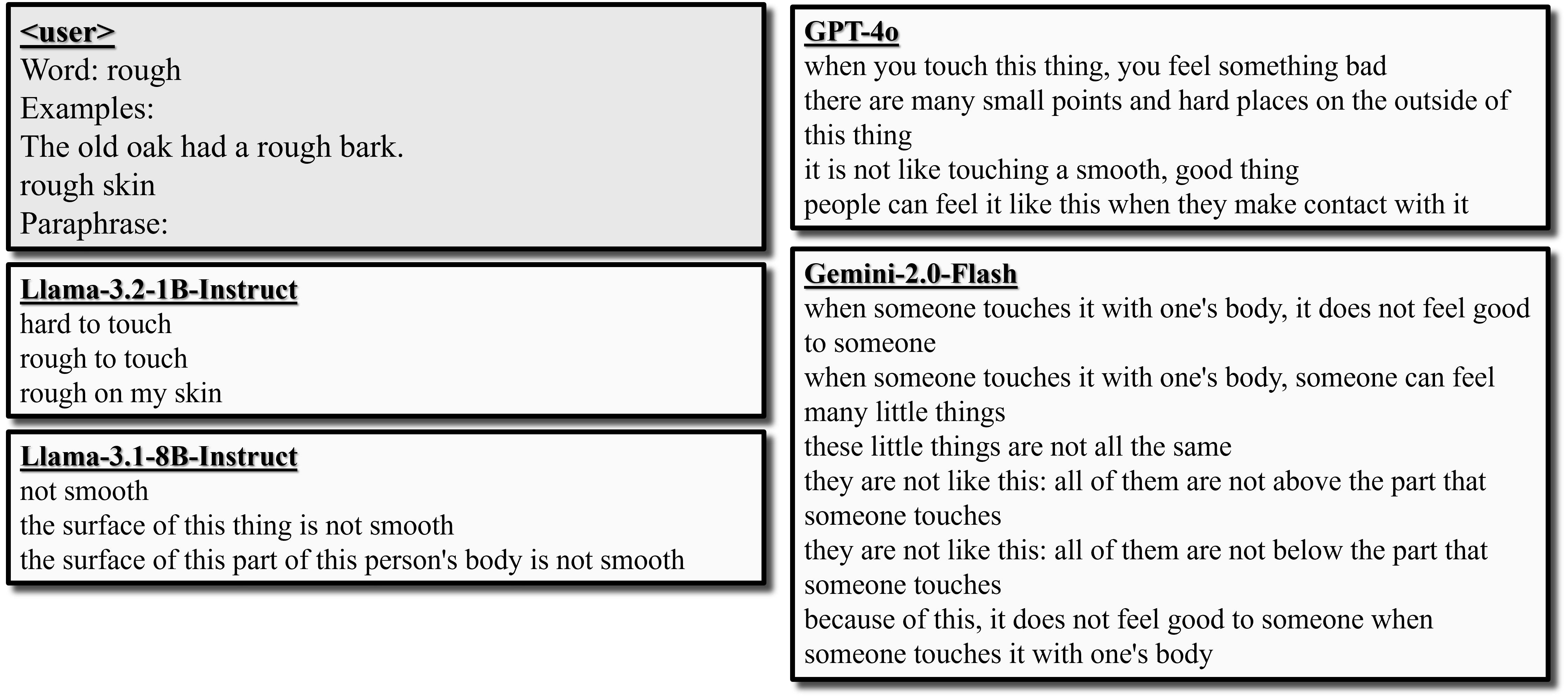}
    \caption{Example outputs from instruction-tuned LLMs when prompted to generate an NSM-style explication for the word "rough." Despite using the same prompt with three few-shot examples, smaller models like Llama-3.2-1B and Llama-3.1-8B fail use the NSM primes in their explications. Larger models such as Gemini 2.0 Flash and GPT-4o produce more structured responses, but their explications still struggle to accurately describe the meaning of the word.}
    \label{fig:example_outputs_rough}
\end{figure}

\paragraph{Lack of Effective and Reliable Models.}
\label{sec:challenges_models}
To evaluate current LLMs’ ability to generate NSM explications, we construct a system prompt (Figure \ref{fig:explication_generation_sys}) based on the task setup in Section \ref{sec:task_setup}, using three few-shot examples drawn from freely available expert-authored NSM explications (Figure \ref{fig:explication_generation_examples}). We apply this prompt to instruction-tuned LLMs of various sizes: Llama-3.2-1B, Llama-3.1-8B, Gemini 2.0-Flash, and GPT-4o. Full prompts and more sample outputs are provided in the Appendix (Sections \ref{sec:example_model_outputs_appendix}, \ref{sec:prompts_appendix}). Figure \ref{fig:example_outputs_rough} presents the responses from each model when prompted to generate an explication for the word "rough." 
As illustrated in the figure and further detailed in Table \ref{tab:experimental_results} (Section \ref{sec:model_experiments}), smaller models such as the Llama variants consistently struggle to generate valid explications using the NSM semantic primes, even after prompt tuning. In contrast, larger models like GPT and Gemini more reliably follow instructions and incorporate the primes in their outputs. However, their paraphrases still frequently fall short of accurately conveying the intended meaning of the target word.
GPT-4o’s explication, while expressing an unpleasant sensation, introduces non-prime terms like “contact” and “points,” which add unnecessary complexity and inaccuracy to its explication, causing it to read more like an explication for "pointy" rather than "rough." Gemini-2.0 Flash attempts to express a similar unpleasant sensation, but its phrasing—such as referencing things “not above/below the part”—introduces confusion, as these spatial qualifiers don’t clearly relate to the sensation of roughness.
The absence of a model that is both effective and efficient for this task highlights a critical gap in using LLMs to generate NSM explications. 

\paragraph{Lack of Datasets.}
\label{sec:challenges_datasets}
Although the absence of effective models creates an opportunity to adapt LLMs for generating NSM explications, progress in that area is prevented by the lack of any datasets for this task. The volume of publicly available, human-authored explications is too limited to support effective fine-tuning, and most existing examples are published under licenses that prevent their inclusion in open datasets. Moreover, these examples would need to be augmented with associated target words and example usages to align with the task setup described in Section \ref{sec:task_setup}. The situation is further complicated by the absence of any standardized benchmarks or evaluation sets, making it difficult to assess or compare future methods consistently.

\paragraph{Lack of Methods for Automatically Evaluating Explication Quality.}
\label{sec:challenges_evaluation}
Critically, the usefulness of any model, dataset, or benchmark is limited without a method for automatically evaluating the quality of generated explications beyond manual analysis. Scalable NSM research requires automated evaluation across three key dimensions: \textit{legality} (correct use of semantic primes and avoidance of circularity), \textit{descriptive accuracy} (how well the explication captures the target word’s meaning), and \textit{cross-translatability} (reliable translation across different languages without distortion of meaning). Developing automated methods for evaluating these aspects is essential for advancing this task. The following section presents the first methods for evaluating NSM explications automatically.

\section{Proposed Methods for Automatic Evaluation of NSM Explications}
\label{sec:evaluation_methods}
In this section, we introduce three automatic evaluation methods for assessing the legality, descriptive accuracy, and cross‑translatability of NSM explications.

\subsection{Explication Legality Scoring}
\label{sec:evaluation_legality}
To evaluate an explication, the first step is to assess how well it follows the principles of the NSM framework: it should rely primarily on semantic primes, minimize the use of semantic molecules (only using them when necessary), and avoid circularity. To implement this, we  count the number of words in the explication that are semantic primes. Any remaining words that are neither semantic primes nor members of the NLTK \cite{bird2006nltk} stopwords list are counted as molecules. We exclude stopwords from being counted as molecules because we observe most serve grammatical functions and contribute little semantic information. To detect circularity, we check if any form of the original word is contained in the explication. 
Next, we define a metric to quantify how “legal” an explication is; that is, how closely it adheres to the NSM framework. We propose the following formula:

\begin{equation}
    \label{eq:legality_score}
    \text{Legality Score} = \frac{\alpha * (\text{primes} - \text{molecules})}{\text{total words in explication}}
\end{equation}

Here, $\alpha$ is a tunable constant weighting the Legality Score in Equation \ref{eq:explicataion_score}, which we set to 10 in this work. A score of 10 represents a “perfectly legal” explication composed entirely of semantic primes, while a score of -10 indicates that no semantic primes are used. We base the calculation on prime/molecule ratios to normalize for differences in explication length. Although this metric does not explicitly account for circularity, circularity is incorporated into the Explication Score metric introduced later in Section \ref{sec:explication_score}.

\subsection{Evaluating Descriptive Accuracy (Substitutability Test)}
\label{sec:substitutability_test}
While the Legality Score captures how many semantic primes and molecules are used in an explication, it does not assess how accurately the explication describes the meaning of the target word. As discussed in Section \ref{sec:nsm_overview}, NSM researchers assess an explication’s descriptive accuracy by substituting it for the target word, ensuring meaning is preserved, and verifying minimality and correct entailments. Our goal is to develop an automated method based on this practice.

(1) We start with selecting an ambiguous passage $x$ containing the target word $w$ masked as \texttt{<UNK>}, ensuring that the target word cannot be easily predicted without additional clues. We then prompt an LLM to predict the masked word and measure the log-probability of the target word’s token(s). Next, we repeat the process, this time providing the explication $e$ for the masked target word (simulating its substitution) and prompting the LLM to predict the masked word. 
We measure the change in log-probability, defined as: 
\begin{equation}
    \Delta_{\text{baseline}} = \log p(w | x, e) - \log p(w | x)
    \label{eq:delta_baseline}
\end{equation}

An effective explication should increase the LLM's likelihood of predicting the correct word compared to when no explication is given, and should yield a larger positive delta compared to a poor explication.

(2) To assess the minimality of an explication, we sequentially remove $k$ lines, one at a time, from the end of the explication, and calculate the average change in log-probability over both removals, defined as: 

\begin{equation}
    \Delta_{\text{min}} = \sum_{i=1}^{k} \log p(w | x,e_{-i}) - \log p(w | x,e_{-i+1})
    \label{eq:delta_min}
\end{equation}

Here, $e_{-i}$ refers to the explication with $i$ lines removed. $e_{-i+1}$ refers to the explication with $i-1$ lines removed, or the full explication if $i=1$. We set $k = 2$, given that explications will almost always contain 3 or more lines.
If an explication is redundant or contains unnecessary details, this change should be small or even positive as lines are removed. In contrast, an appropriately minimal explication should show a negative delta, indicating a loss of meaningful information in the explication with each removal. 

(3) To test whether an explication captures the target word’s entailments, we sequentially remove $k$ sentences, one at a time, from the passage (excluding the one with \texttt{<UNK>}) and measure the average change in log-probability, defined as: 

\begin{equation}
    \Delta_{\text{ent}} = \sum_{j=1}^{k} \log p(w | x_{-j},e) - \log p(w | x_{-j+1},e)
    \label{eq:delta_ent}
\end{equation}

We use the same $k=2$ from the minimality tests for consistency. Here, $x_{-j}$ refers to the explication with $j$ lines removed. $x_{-j+1}$ refers to the passage with $j-1$ lines removed, or the full passage if $j=1$.
If the model’s prediction remains stable or improves, it suggests the explication contains the core meaning needed to infer the word. A drop in probability implies the model relied on information from the removed context—entailments the explication failed to supply. 
For example, if removing “She was feeling very sad” from the passage lowers the model’s confidence in predicting "cry," the explication likely missed conveying that emotional state. 

For an explication, we perform these tests repeatedly over multiple ambiguous passages $P$ and LLMs $G$, averaging the outcomes from all of them in order to generalize the result and reduce sensitivity to any particular passage or model behavior. The LLMs we use in this process include instruction-tuned Llama-3.1-8B, Mistral-7B, and Gemini-3-12B. Finally, we calculate a "substitutability score" to assess the descriptive accuracy of an NSM explication, with the following formula:

\begin{equation}
\label{eq:average_substitutability_score}
\text{Substitutability Score} = \frac{1}{|G||P|} \sum_{g \in G} \sum_{p \in P} \left( \min(\beta, \Delta^{(g,p)}_{\text{baseline}} - \Delta^{(g,p)}_{\text{min}} + \Delta^{(g,p)}_{\text{ent}}) \right)
\end{equation}

This scoring formula rewards explications that are descriptively accurate (improving target word prediction), minimal, and have the right entailments. $\beta$ is a cap that can be set to limit the maximum substitutability score, preventing extreme values from skewing the evaluation. 
Before running evaluations, we verified that the LLMs we used for the subsitutability testing produce log-probability values on a comparable scale, as shown in Figure \ref{fig:score_distribution} in the Appendix. We observed that the models, particularly Gemma, can sometimes assign unusually large log-probabilities. A manual review of these instances revealed that these extreme values were likely due to idiosyncrasies, rather than indicative of a particularly accurate explication. As a result, we decided to set $\beta$ to 40; this also helps keep the overall explication score (Section \ref{sec:explication_score}) within a 100-point scale.
To our knowledge, this is the first automated method for evaluating NSM explication accuracy without requiring human input. We include pseudocode for this process in Section \ref{algo:substitutability_score} of the Appendix.

\subsection{Overall Explication Score}
\label{sec:explication_score}
To provide a metric that can holistically assess both the legality and descriptive accuracy of an explication, we compute an overall explication score by combining the above two metrics:

\begin{equation}
    \label{eq:explicataion_score}
    \text{Explication Score} = \gamma * (\text{Substitutability Score} + \text{Legality Score})
\end{equation}

We set $\gamma$, a tuneable constant, to 2 to normalize the max score to 100, with the substitutability score contributing up to 40 points and the legality score ranging from -10 to 10. Circular explications (containing the target word) are automatically scored as 0. Future work may explore alternative settings of $\alpha, \beta, k,$ and $\gamma$.

\begin{figure}
    \centering
    \includegraphics[width=.9\linewidth]{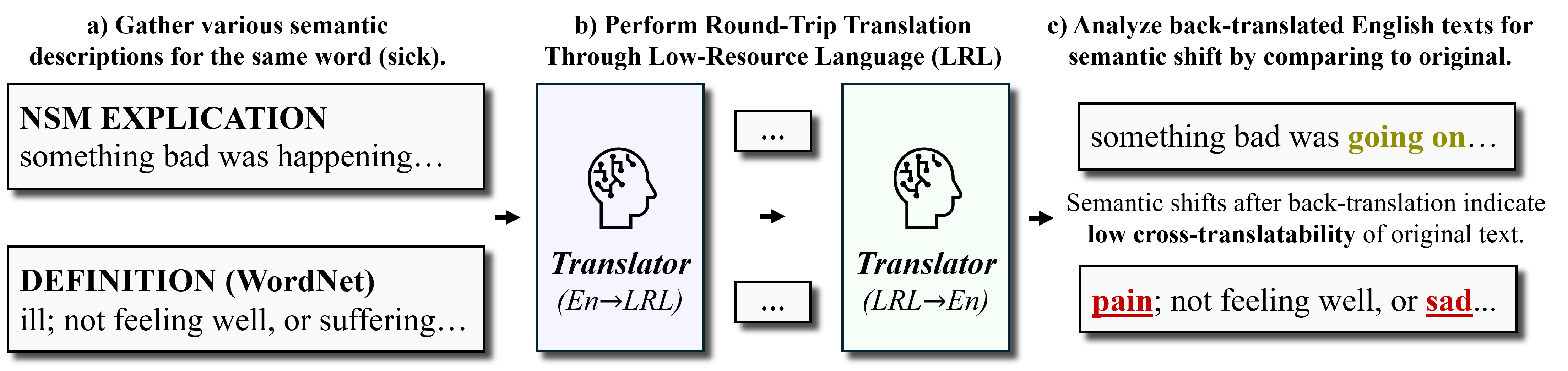}
    \caption{Illustration of the cross-translatability test described in Section \ref{sec:cross_translate_eval}. Semantic descriptions are translated into a low-resource language supported by Google Translate and then back into English. Any semantic drift in the back-translated text indicates difficulty in consistent translation, suggesting the texts contains semantically complex, hard to translate words.}
    \label{fig:cross_translatability_test}
\end{figure}

\subsection{Evaluating Cross-Translatability}
\label{sec:cross_translate_eval}
A core requirement of NSM explications is that they be easily cross-translatable. To test the cross-translatability of an explication (or any text), we propose the procedure illustrated in Figure \ref{fig:cross_translatability_test}. We collect various semantic descriptions for a word, such as NSM explications and dictionary definitions. Using Google Translate as our translator, we focus on low-resource languages with known lower translation quality, in order to better stress-test the text. The semantic descriptions in English text are translated into the target language and then back into English. We assess back-translated English texts by comparing them to their originals, using BLEU scores and embedding-based similarity to measure lexical and semantic differences. Lower scores on either metric indicate greater semantic drift, suggesting more difficulty in consistent translation in and out of the target language. Although NSM explications are longer than dictionary definitions, they are expected to perform better due to their use of semantic primes, which should reduce the semantic complexity of the translation task.

\section{Proposed Dataset}
\label{sec:dataset}
In this section, we introduce the first dataset designed explicitly for the task of NSM explication generation, which serves as the basis for fine-tuning the DeepNSM model described in Section \ref{sec:model_experiments}. Following the task formulation outlined in Section \ref{sec:task_setup}, each entry in the dataset includes a target word, several example sentences illustrating its usage, and a corresponding NSM explication.

\begin{figure}
    \centering
    \includegraphics[width=.9\linewidth]{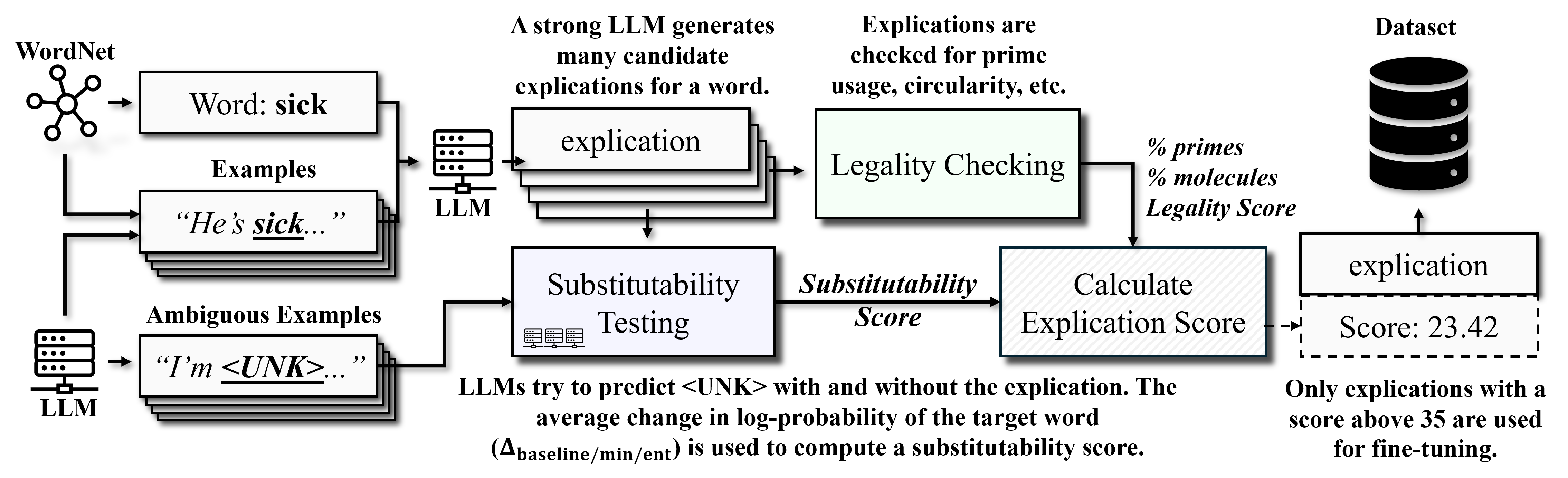}
    \caption{Our dataset generation process described in Section \ref{sec:dataset_generation_framework}. Target words are selected from WordNet and paired with example usage sentences. Gemini-2.0-Flash generates many candidate explications for the word, and the evaluation methods outlined in Section \ref{sec:evaluation_methods} are used to select the highest quality explications for the final dataset.}
    \label{fig:dataset_framework}
\end{figure}

\textbf{Dataset Generation.}
\label{sec:dataset_generation_framework}
To ensure our dataset covers a wide range of word-meanings, we use WordNet \cite{miller1995wordnet}, a lexical database that provides structured definitions for various word senses, grouped into synsets. Synsets are sets of synonymous words that share a single sense or meaning. After filtering out NSM primes
, we obtain 88,078 unique word senses. While WordNet includes some example sentences, many synsets lack examples, and none offer ambiguous passages which could be used for the substitutability tests we define in Section \ref{sec:substitutability_test}. Figure \ref{fig:dataset_framework} demonstrates our dataset generation process. We first select a target word sense from a WordNet synset, applying filters to exclude NSM primes. To address gaps in example sentences, we use instruction-tuned LLMs (Llama-3-1/8B, Mistral-7B, Gemma-2-2/9B, Llama-3.2-3B, Gemma-2-3B) to generate additional examples, prompted with the WordNet definition to ensure the examples align with the word sense.

For each word sense, we gather 2 to 5 example sentences using the target word in context. We then prompt an LLM to generate multiple candidate NSM explications based on the target word and its examples. We select Gemini-2.0-Flash for this step, as it offers a good balance between explication quality and speed. With a temperature setting of 0.7, Gemini generates over 440,000 candidate explications across all synsets. As noted in Section \ref{sec:challenges}, the output quality of general-purpose LLMs can vary significantly. To address this, we use the evaluation metrics from Section \ref{sec:evaluation_methods} to identify and remove low-quality generations. The filtered dataset aims to support fine-tuning more effective and consistent LLMs for NSM explication generation.

Once candidate explications are generated, we first perform legality checks on each and calculate a legality score (Equation \ref{eq:legality_score}). Then, we assess descriptive accuracy through substitutability testing, generating four ambiguous example paragraphs with an LLM. We compute a substitutability score for each explication using three grader LLMs (Llama-3.1-8B, Mistral-7B-v0.3, and Gemma-3-12B), following Section \ref{sec:substitutability_test}. Legality and substitutability scores are then combined to compute the overall Explication Score (Equation \ref{eq:explicataion_score}). This score serves as a quality indicator that can be used to filter out low-quality entries from the final dataset. Finally, the word, scored explication, and example sentences are added as a candidate entry in the dataset.

\textbf{Evaluation Set/Benchmark.}
\label{sec:dataset_evaluation_split}
For the evaluation split, we hand-curated 149 unique words across various semantic categories, such as nouns, mental predicates, and speech act verbs. Each word was mapped to its WordNet synsets to ensure that neither the word, its synonyms, nor alternative senses from the training data were included, preventing contamination. We selected two to five usage examples and four ambiguous paragraphs for consistent substitutability testing, with examples and contexts manually refined for clarity and coherence.

\textbf{Quality Filtering and Final Dataset.}
\label{sec:dataset_quality_filtering_final}
After generating and scoring candidate explications, we filter out those with scores below 35, a quality threshold met by fewer than 15\% of candidates, as most scores range from 10 to 20.
To prevent overrepresentation, we cap explications per word sense at two. The final dataset contains around 44,000 word-example-explication entries, with 1,000 set aside for validation and 43,000 for training. The hand-curated test set of 149 entries is used for evaluation. 

\begin{table}
  \caption{Evaluation of NSM explications generated by LLMs for the benchmark set introduced in Section \ref{sec:dataset_evaluation_split}. 
  An up (down) arrow for a metric means higher (lower) is better. "Dictionary Def." are existing definitions provided from WordNet. 
  \underline{Best underlined}.
  DeepNSM models surpass SOTA general LLMs for NSM explication generation despite only having 1B and 8B parameters.
  }
  \label{tab:experimental_results}
  \centering
  \begin{tabular}{lcccccc}
  \toprule 
  \multicolumn{1}{c}{Model} & 
  \makecell{\textbf{Explication} \\ \textbf{Score} $\uparrow$}  & 
  \makecell{Legality \\ Score $\uparrow$}  & 
  \makecell{Substitutability \\ Score $\uparrow$}  & 
  \makecell{Primes \\ Ratio $\uparrow$}  & 
  \makecell{Molecules \\ Ratio $\downarrow$}  & 
  \makecell{Circular \\ \% $\downarrow$}  \\
  \midrule
  Dictionary Defs. & \textbf{13.4} & -4.7 & \underline{12.14} & 8.0 & 55.1 & 10.0 \\
  \midrule
  Llama-3.2-1B-it & \textbf{6.1} & -0.4 & 6.82 & 29.2 & 33.2 & 45.6 \\
  Llama-3.1-8B-it & \textbf{20.0} & 2.3 & 7.86 & 43.9 & 20.9 & 2.0 \\
  Gemini-2.0-Flash & \textbf{22.2} & 5.1 & 6.40 & 62.1 & 11.5 & 2.7 \\
  GPT-4o & \textbf{22.9} & 4.6 & 6.95 & 59.9 & 13.8 & \underline{1.3} \\
  \midrule
  DeepNSM-1B$^\dagger$ & \textbf{19.7} & 5.0 & 5.22 & 61.1 & 10.9 & 4.7 \\
  DeepNSM-8B$^\dagger$ & \textbf{22.2} & 4.9 & 6.40 & 60.0 & 11.5 & \underline{1.3} \\
  DeepNSM-1B & \textbf{23.2} & 5.1 & 7.02 & 61.9 & 10.6 & 5.4 \\
  DeepNSM-8B & \textbf{\underline{24.6}} & \underline{5.4} & 7.34 & \underline{63.9} & \underline{10.4} & 3.3 \\
  \bottomrule
\end{tabular}
\raggedright
{\footnotesize $^\dagger$Trained on a dataset with no quality filtering applied.}
\vspace{-1em}
\end{table}

\section{DeepNSM Model and Experiments}
\label{sec:model_experiments}

In this section, we present DeepNSM, the first LLM fine-tuned specifically for generating NSM explications. We evaluate DeepNSM against several baseline LLMs and demonstrate that fine-tuning on our dataset introduced in Section \ref{sec:dataset} enables DeepNSM to produce NSM explications that are both higher in quality and more easily translatable across languages.

\subsection{Setup}
\label{sec:setup}
To address key model limitations described in Section \ref{sec:challenges}, we fine-tune Llama 3.2 1B and Llama 3.1 8B models using our curated dataset. All fine-tuning runs are conducted for one epoch on an NVIDIA H100 GPU. Full training details can be found in Section \ref{sec:fine_tuning_appendix}. We evaluate all models using the evaluation set described in Section \ref{sec:dataset_evaluation_split}. For each target word in the benchmark, models are prompted to generate an explication given the word and its examples. All baseline models are prompted with a consistent system message and a fixed set of few-shot examples. The generated explications are evaluated using the legality, substitutability, and cross-translatability tests described in Section \ref{sec:evaluation_methods}, as well as qualitative human ranking. We compare DeepNSM models (1B and 8B) with the following instruction-tuned LLM baselines: Llama-3.2-1B, Llama-3.1-8B, Gemini-2.0-Flash (estimated $\sim$70B parameters), and GPT-4o (200B estimated \cite{abacha2025medecbenchmarkmedicalerror}). We also include the existing WordNet definitions (“Dictionary Def.”) as a non-LLM, non-NSM baseline. Similar to fine-tuning, all evaluations are conducted on an NVIDIA H100 GPU. Error bars for all experiments can be viewed in Tables \ref{tab:experimental_results_error_bars}, \ref{tab:crosslingual_bleu}, and \ref{tab:crosslingual_embed} in the Appendix.

\begin{table}
\caption{Cross-translatability results following the method described in \ref{sec:cross_translate_eval}. We test on model-generated explications for five low-resource languages from different families: Alur (alz), Kinyarwanda (rw), Dzongkha (dn), Dinka (din), and Abkhaz (ab). After performing round-trip translation, we measure the semantic shift of the back-translated text using BLEU and Embedding Similarity. \underline{Best underlined}. NSM explications consistently show less semantic drift, reflecting their universality.
}
\centering
\begin{tabular}{l|ccccc|ccccc}
\toprule
 & \multicolumn{5}{c|}{BLEU $\uparrow$} & \multicolumn{5}{c}{Embedding Similarity $\uparrow$} \\
\multicolumn{1}{c|}{Model} & alz & rw & dn & din & ab & alz & rw & dz & din & ab \\
\midrule
Dictionary Def. & 22.9 & 35.1 & 19.6 & 24.4 & 26.9 & 69.6 & 79.6 & 73.3 & 69.1 & 78.0 \\
\midrule
Llama-3.2-1B-it & 29.0 & 45.0 & 21.0 & 25.8 & 25.2 & 79.3 & 88.0 & 81.3 & 77.3 & 84.2 \\
Llama-3.1-8B-it & 33.2 & 50.8 & \underline{23.8} & 31.8 & 27.8 & 81.6 & 89.8 & 81.8 & 78.1 & 84.0 \\
Gemini-2.0-Flash & 33.5 & 56.3 & 23.2 & 32.2 & 30.1 & 84.0 & \underline{93.0} & 81.7 & 78.5 & 85.1 \\
GPT-4o & 31.2 & 51.3 & 21.6 & 31.2 & 28.7 & 82.0 & 91.4 & 78.9 & 77.4 & 84.8 \\
\midrule
DeepNSM-1B & 34.5 & \underline{56.9} & 23.6 & \underline{32.6} & 31.9 & 81.6 & 91.4 & 80.4 & 78.0 & 82.7 \\
DeepNSM-8B & \underline{37.0} & 55.6 & 23.3 & 32.4 & \underline{33.2} & \underline{84.2} & 91.7 & \underline{82.0} & \underline{79.0} & \underline{85.4} \\
\bottomrule
\end{tabular}
\label{tab:crosslingual_eval}
\end{table}

\subsection{Results}
\label{sec:results}

\paragraph{DeepNSM Enables High-Quality, Efficient NSM Explication Generation.} 
Table \ref{tab:experimental_results} shows that despite having only 1B and 8B parameters, all DeepNSM variants consistently outperform or match state-of-the-art general-purpose models across key metrics, including legality, substitutability, prime usage, molecule usage, and circularity. Notably, DeepNSM models achieve the highest overall explication score, with even the 1B model outperforming all other models, regardless of size.
While dictionary definitions and Llama baselines score higher on substitutability, these results are misleading because they rely on poor prime usage and high circularity, often reusing the target word, effectively “cheating” the NSM paraphrasing task. In contrast, the fine-tuned DeepNSM 1B and 8B models improve prime usage by 30.9\% and 16.1\% and reduce molecule usage by 22.3\% and 9.4\%, respectively. These findings demonstrate that high-quality NSM explication generation is achievable in smaller models with our proposed dataset, addressing the model limitations discussed in Section \ref{sec:challenges} without requiring large-scale compute or commercial APIs.

\paragraph{Dataset Quality Filtering Improves Explication Accuracy and Prime Usage.}
To isolate the impact of dataset quality filtering, we fine-tune two versions of each DeepNSM model: one on the high-quality, filtered dataset (Explication Score $\geq$ 35) described in Section \ref{sec:dataset_quality_filtering_final}, and another on a randomly sampled, unfiltered dataset of the same size (marked with $\dagger$), drawn from the full pool of candidate explications without regard to quality. Results in Table \ref{tab:experimental_results} show that, compared to their unfiltered counterparts, the DeepNSM 1B and 8B models trained on quality-filtered data achieve a 34\% and 15\% relative increase in substitutability scores, respectively. Prime usage also improves, with the DeepNSM 8B model showing a 3.9 percentage point gain. These improvements demonstrate the effectiveness of our filtering strategy in enhancing both descriptive accuracy and semantic prime usage. By producing the first high-quality dataset for NSM explication generation, our work directly addresses the dataset limitations outlined in Section \ref{sec:challenges}.

\paragraph{NSM Explications Demonstrate High Cross-Translatability for Low-Resource Languages.}
We perform cross-translatability testing (Section \ref{sec:cross_translate_eval}) on model-generated explications with five low-resource languages: Alur, Dinka, Kinyarwanda, Dzongkha, and Abkhaz. We then measure the match between the back-translated explication and the original using BLEU and embedding similarity, with embeddings generated from BERT-based sentence embeddings in the Sentence-Transformers library \cite{reimers2019sentencebert}. Table \ref{tab:crosslingual_eval} shows that NSM explications are significantly more resistant to semantic drift during translation to and from low-resource languages compared to dictionary-style definitions. This is particularly evident in Kinyarwanda, where NSM explications outperform by over 20 BLEU points and 10 embedding similarity points. Although we did not explicitly use cross-translatability metrics during dataset generation, explications produced by DeepNSM consistently achieve the highest cross-translatability scores across all languages tested. This indicates that our dataset improves both the descriptive accuracy of model-generated explications and their cross-translatability into different languages.
These results suggest that DeepNSM could already serve as a tool for helping rephrase English texts to make them more accessible and readily translatable for speakers of many languages.

\paragraph{Metrics Align with Qualitative Judgements.}
We also conduct a qualitative analysis to evaluate the alignment between our automatic metrics and human judgments. To do this, we perform a blind review and ranking of the explications generated by all models evaluated in Table \ref{tab:experimental_results}. Explications from DeepNSM models received top rankings 46\% of the time, with 33\% for the 8B model and 13\% for the 1B model, which frequently placed second to 8B. This was significantly higher than the top rankings received by GPT (28\%), Gemini (21\%), and the Llama models (5\%). These findings indicate that automatic metrics align with human qualitative analysis and highlight the effectiveness of the dataset filtering strategy, as DeepNSM trained on filtered data consistently outperforms other models. These results support the effectiveness of our metrics in addressing the critical challenge outlined in Section \ref{sec:challenges}, providing the first automated evaluation method for NSM explications.

\section{Related Work}
\label{sec:related_work}
Prior work has applied NSM to computational tasks, such as encoding NSM texts in PROLOG \cite{zamblera_computational_nodate}, creating word vectors aligned with semantic primes
for the ACT-R cognitive architecture \cite{rodgers_creating_nodate}, and using NSM for building semantic graphs using marker-passing algorithms \cite{fahndrich_semantic_nodate}. However, these methods do not use deep learning or target LLMs, and none address automatic explication generation, which is the central focus of our work.
\cite{giulianelli_interpretable_2023} use LLMs to study semantic drift by generating contextualized definitions from target words and usage examples. However, their work does not target NSM, and thus these definitions can be inherently more prone to circularity and poor cross-translatability. Other work \cite{periti_lexical_2024} tracks semantic drift through changes in LLM-derived embeddings. Incorporating NSM into such analyses is an promising research direction which can be catalyzed by this work.
Multilingual embeddings \cite{chen2024bge, conneau2019unsupervised} project semantically related words from different languages into a shared space to support cross-lingual tasks. Yet, they often lack the semantic precision of English-focused LLMs and depend heavily on large multilingual corpora. To enhance generalization, \cite{man_lusifer_2025} align a multilingual encoder with an English-centric LLM to create a "universal embedding space." However, this space is purely learned and lacks grounding in any semantic framework. Such work highlights broader interest for universal semantic representations in LLMs and the relevance of this work.

\section{Conclusion}
\label{sec:conclusion}
This work introduces the first use of LLMs for the NSM approach to semantic analysis, along with the first models, dataset, and evaluation methods, laying the groundwork for future research. The potential impact is significant, as we believe AI can scale the NSM approach to reveal entire languages as vast articulations of semantic primes, enabling new possibilities for cross-linguistic and cross-cultural understanding, as well as promising applications for many tasks. This marks a significant step towards using AI to bridge human languages through their universal semantic core. 


\label{sec:limitations}
\textbf{Limitations.} This work, as the first to adapt LLMs to the NSM approach, lays the foundation for methods and evaluation protocols. However, two areas remain unexplored. First, we do not explore techniques such as modified generation or restricting logits to semantic primes, nor do we examine incorporating proposed grammatical constraints on prime combinations \cite{goddard_1_2008}, which could further improve structural alignment with the NSM framework in LLM-generated explications. Second, our substitutability evaluations rely on 7–12B parameter grader models and exclude encoder-only architectures like BERT; further investigation into different grader models may improve alignment with human judgments. 
Finally, the DeepNSM outputs and the explications in our dataset should not be considered fully accurate representations of word meanings, and LLM-generated explications should be human-verified before use in authoritative contexts. 

\bibliography{custom}
\bibliographystyle{abbrv}

\newpage

\appendix

\section*{\LARGE Appendix}
This appendix offers supplementary materials, background information, and extended discussions that could not be included in the main paper due to space constraints.

\section{Semantic Primes}
\label{sec:semantic_primes_appendix}
Semantic primes are treated as irreducible semantic units that cannot be defined using simpler terms without leading to circularity, where either the word itself appears in its own definition, or the words used to define it lead back to the original word. These meanings are taken to be self-evident and universally present across languages. A particular language’s translation of the semantic primes are often referred to as exponents. Importantly, primes can display polysemy in specific languages, potentially leading to misinterpretation \cite{goddard_semantic_2014, goddard_1_2008, goddard_semantic_2010}. For instance, the English exponent “above” is a valid prime in the spatial sense (“the sky is above the ground”), but not in the metaphorical sense (“I am above this work”). Additionally, primes may exhibit allolexy, that is, they may be represented by multiple synonymous words in a language \cite{goddard_1_2008, goddard_semantic_2010, goddard_semantic_2014}. Semantic primes have also been proposed to have associated grammatical properties \cite{goddard_1_2008}.

\section{Methodological Advantages of NSM}
\begin{figure}[h!]
    \centering
    \includegraphics[width=0.9\linewidth]{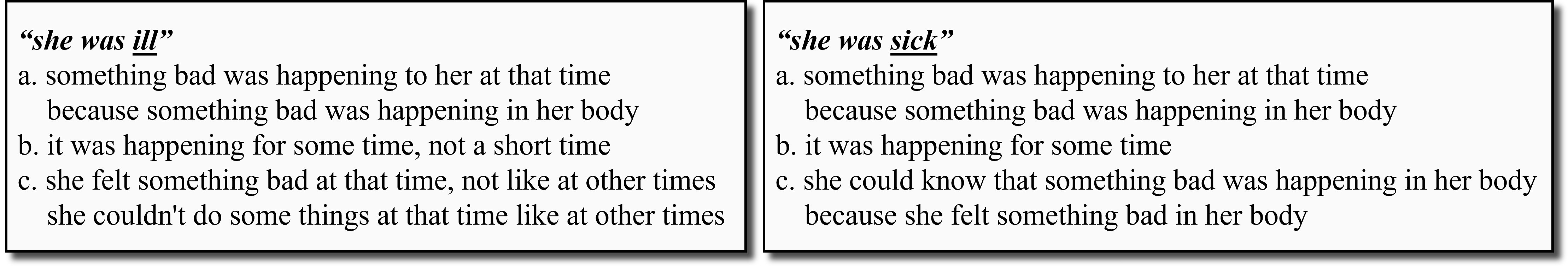}
    \caption{Comparing the meaning of "ill" and "sick" using their NSM explications. Explications are taken from \cite{goddard_semantics_2021}. }
    \label{fig:ill_vs_sick_comparison}
\end{figure}

\footnote{This section is largely paraphrased from \cite{goddard_1_2008}.}
\label{sec:methodological_advantages_nsm}
To support testable predictions about real language use, a semantic description requires a clear and transparent link between the semantic description and the natural language it aims to explain. When that link becomes too abstract or detached, empirical validation can become more challenging. For instance, the common formal analysis of "x kills y" as [\texttt{CAUSE} x (\texttt{DIE} y)] fails to account for usage differences between "kill" and "cause to die," yet rather than revising the analysis, some have argued that \texttt{CAUSE} doesn’t mean the same as the English word cause \cite{mccawley1972program}. However, such a move can undermine empirical testing by insulating the analysis from being supported by real-world evidence. While formal semanticists and logicians may see the ordinary natural language of NSM as too imprecise for semantic representation, as \cite{allan1986linguistic} argues, even trained experts inevitably rely on their native languages when learning formal systems and when discussing the intuitions behind their formalisms.
Moreover, NSM paraphrases still preserve all the advantages of a highly disciplined system for semantic analysis. By limiting paraphrases to semantic primes, the framework enables direct and structured comparisons between word meanings. For example, we can compare the words "ill" and "sick" via their explications, shown in Figure \ref{fig:ill_vs_sick_comparison}.

A close comparison of these explications reveals a key distinction: in line $b$, the explication for ill includes the additional words “not a short time.” This suggests that ill is typically used to describe longer-term conditions, while sick may apply to shorter or even momentary states of discomfort. From this difference, we can directly formulate testable predictions about how these words are used. For instance, if ill typically refers to something that lasts "not a short time," we would expect it to occur less frequently in contexts containing the semantic prime "a short time" or other words that, according to their explications, happen for "a short time." If semantic descriptions are restricted to combinations of a finite set of semantic primes, then generating and testing these kinds of predictions about real-world language use, such as differences in word usage or co-occurrence patterns, can be systematically automated, i.e., testing for the aforementioned collocations whenever "not a short time" appears in an explication.

Another methodological benefit of restricting semantic analysis to a set of universal semantic primes is that it helps avoid presumptively imposing foreign cultural or terminological categories onto the languages being studied. This is especially important when working with less widely spoken or native languages, where introducing a formal metalanguage may impose a semantic component like \texttt{CAUSE} onto a language where the word "cause" has no corresponding verb \cite{goddard1991testing}. Such impositions risk misrepresenting the language's conceptual distinctions and culturally specific meanings. By relying solely on universal semantic primes for analysis, the NSM approach ensures that semantic descriptions remain free from any culturally specific vocabulary and thus broadly cross-translatable.

\section{Psuedocode for Computing Substitutability Score}
\begin{algorithm}
\label{algo:substitutability_score}
\caption{Compute Substitutability Score (Generalized for any $k$)}
\begin{algorithmic}[1]
\Require Target word $w$, sense definition $s$, explication $e$, grader LLMs $\{G_i\}$, truncation depth $k$, cap $\beta$
\State Generate ambiguous passages $\{P_j\}$ using GPT-4o mini with $w$ and $s$, masking $w$ as \texttt{<UNK>}

\For{each grader LLM $G_i$}
    \For{each passage $P_j$}
        \State $x_0 \gets P_j$ with \texttt{<UNK>} masking $w$
        \State $\Delta_{\text{baseline}}^{(i,j)} \gets \log p(w \mid x_0, e) - \log p(w \mid x_0)$

        \Comment{Compute $\Delta_{\text{min}}$}
        \State Initialize $\Delta_{\text{min}}^{(i,j)} \gets 0$
        \For{$m = 1$ to $k$}
            \State $e_{-m+1} \gets$ explication with $(m-1)$ lines removed from the end
            \State $e_{-m} \gets$ explication with $m$ lines removed from the end
            \State $\Delta_{\text{min}}^{(i,j)} \mathrel{+}= \log p(w \mid x_0, e_{-m}) - \log p(w \mid x_0, e_{-m+1})$
        \EndFor
        \State $\Delta_{\text{min}}^{(i,j)} \gets \Delta_{\text{min}}^{(i,j)} / k$

        \Comment{Compute $\Delta_{\text{ent}}$}
        \State Initialize $\Delta_{\text{ent}}^{(i,j)} \gets 0$
        \For{$m = 1$ to $k$}
            \State $x_{-m+1} \gets$ passage with $(m-1)$ non-\texttt{<UNK>} lines removed (excluding the \texttt{<UNK>} line)
            \State $x_{-m} \gets$ passage with $m$ non-\texttt{<UNK>} lines removed (excluding the \texttt{<UNK>} line)
            \State $\Delta_{\text{ent}}^{(i,j)} \mathrel{+}= \log p(w \mid x_{-m}, e) - \log p(w \mid x_{-m+1}, e)$
        \EndFor
        \State $\Delta_{\text{ent}}^{(i,j)} \gets \Delta_{\text{ent}}^{(i,j)} / k$

        \State $\text{score}_{ij} \gets \min(\beta, \Delta_{\text{baseline}}^{(i,j)} - \Delta_{\text{min}}^{(i,j)} + \Delta_{\text{ent}}^{(i,j)})$
    \EndFor

    \State $\text{score}_i \gets \text{mean}_j(\text{score}_{ij})$
\EndFor

\State \Return Average substitutability score: $\frac{1}{|G|} \sum_i \text{score}_i$
\end{algorithmic}
\end{algorithm}


\begin{figure}
    \centering
    \includegraphics[width=1\linewidth]{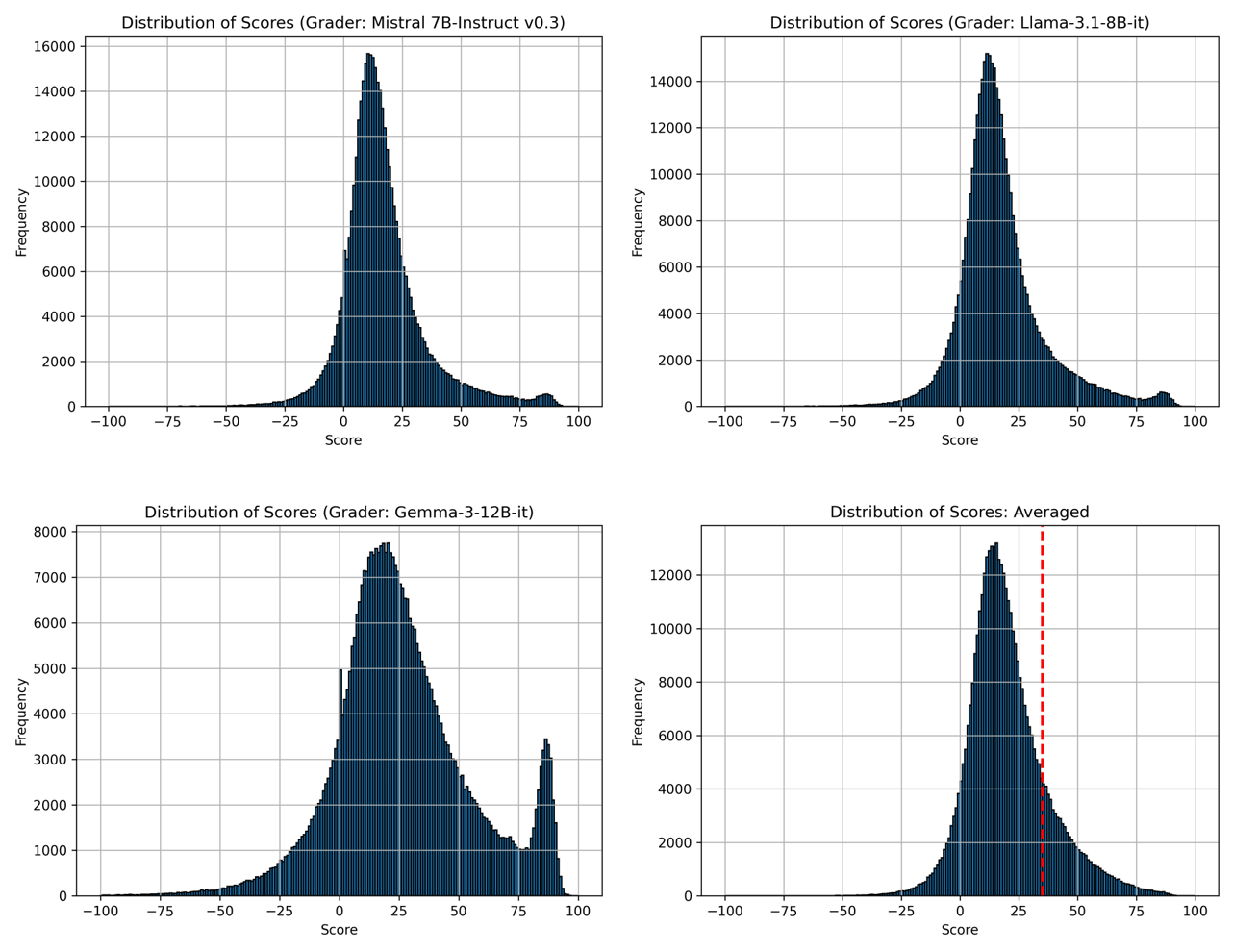}
    \caption{Distribution of Explication Scores assigned by different "grader" LLMs during substitutability testing, evaluating all candidate explications generated through the dataset creation process described in Section \ref{sec:dataset_generation_framework}. The bottom right plot shows the distribution of scores after averaging the scores given by all graders. The red line indicates the cutoff threshold of $\geq$ 35 from Section \ref{sec:dataset_quality_filtering_final}.}
    \label{fig:score_distribution}
\end{figure}

\section{Further Details for Dataset}
\label{sec:dataset_appendix}
As described in Section \ref{sec:challenges_datasets}, there are no publicly available datasets of NSM explications, and too few human-written examples to build one from existing sources. To address this, we construct a dataset using large language models and existing lexical resources, applying the evaluation methods from Section \ref{sec:evaluation_methods} to ensure quality. We fully detail the dataset construction process below.

\subsection{Dataset Structure}
\label{sec:dataset_structure_appendix}
Following the task setup in Section \ref{sec:task_setup}, each dataset entry consists of a target word, two to five example sentences showing how the word is used, and an associated NSM explication for that word.

\subsection{Dataset Source}
\label{sec:dataset_source_appendix}
To ensure our dataset captures a broad range of word meanings, we use WordNet \cite{miller1995wordnet}, a lexical database that organizes words into synsets, groups of synonyms that share a common meaning. WordNet also provides a dictionary-style definition for each word sense and, in some cases, includes example sentences demonstrating its usage, though these examples are relatively uncommon. WordNet is particularly useful for capturing polysemy, since different words can belong to the same synset. To build our base vocabulary for dataset generation, we iterate over all individual words in WordNet, rather than just synsets. This allows us to explicitly account for polysemous words with multiple senses. We filter out NSM primes during this process. After filtering, we are left with 88,078 unique word senses.

\subsection{Example Generation}
\label{sec:example_generation_appendix}
As noted in Section \ref{sec:dataset}, most words in WordNet either lack example sentences or include only one, while our goal is to provide 2–5 examples per word. To address this, we use instruction-tuned LLMs (Llama-3-1/8B, Mistral-7B, Gemma-2-2/9B, Llama-3.2-3B, and Gemma-2-3B) to generate additional example sentences. Each model is prompted with the WordNet definition to ensure the generated examples match the intended word sense. We apply this process to all 88,078 filtered word senses from WordNet. After generation, each word sense is associated with over 20 example sentences, from which we randomly select 2-5 when constructing the dataset.

\begin{figure}[h!]
    \centering
    \includegraphics[width=1\linewidth]{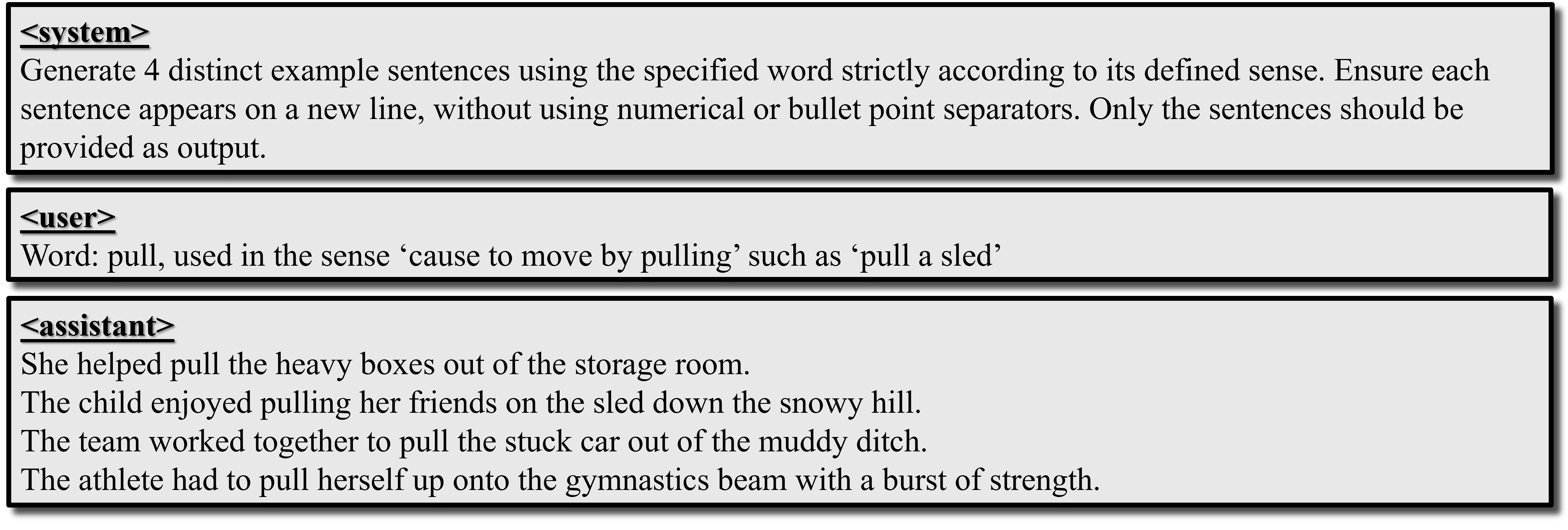}
    \caption{Prompt for generating additional usage examples.}
    \label{fig:example_generation}
\end{figure}

\begin{figure}[h!]
    \centering
    \includegraphics[width=1\linewidth]{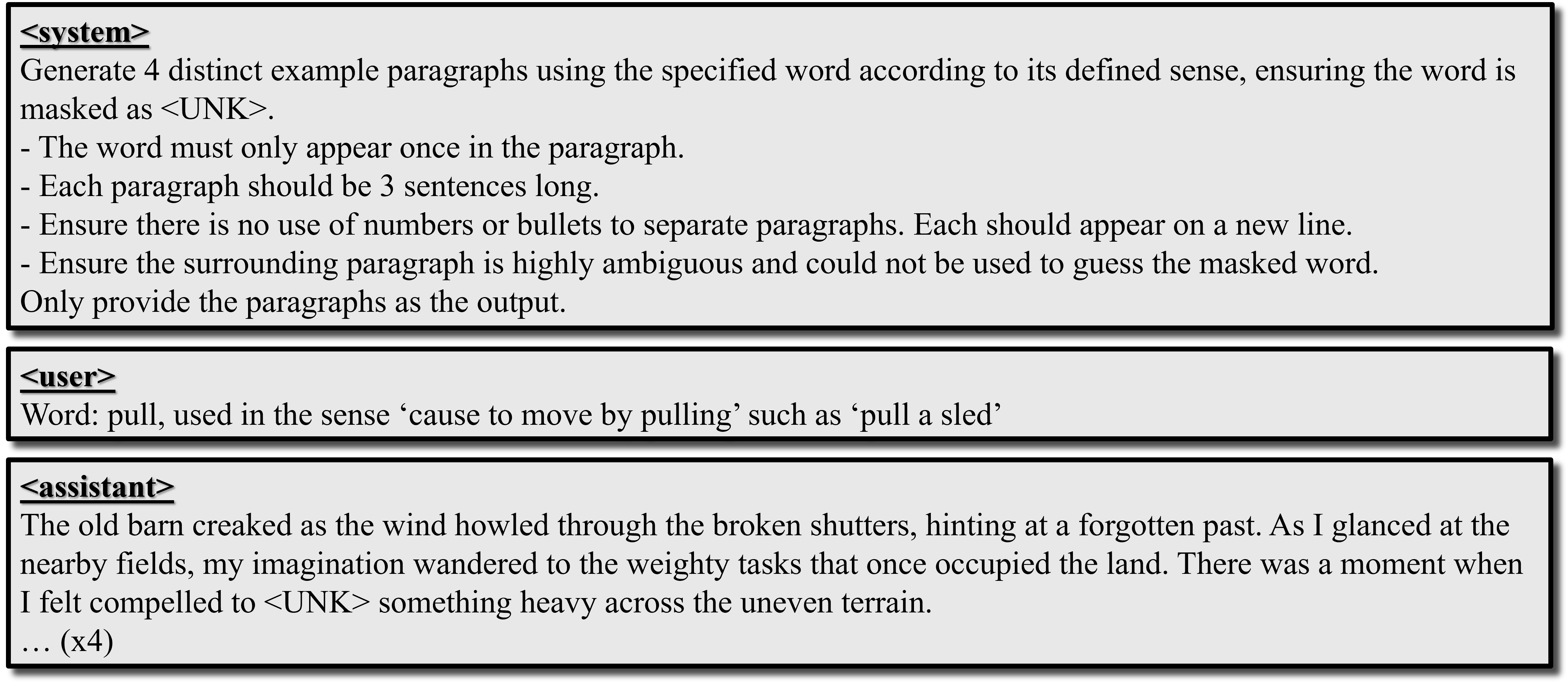}
    \caption{Prompt for generating additional ambiguous examples for the substitutability testing described in Section \ref{sec:substitutability_test}.}
    \label{fig:example_generation_ambig}
\end{figure}

WordNet also does not provide ambiguous example passages that can be used for the substitutability testing described in Section \ref{sec:substitutability_test}. To address this, we use GPT-4o mini to generate four ambiguous example paragraphs for each target word, formatted specifically to support the substitutability evaluation procedure. Figure \ref{fig:example_generation} and Figure \ref{fig:example_generation_ambig} show the prompts used for example generation.

\section{DeepNSM Model and Experiments}
\label{sec:model_experiments_appendix}
We include the experimental results from the main paper with error bars in Table \ref{tab:experimental_results_error_bars}, Table \ref{tab:crosslingual_bleu}, and Table \ref{tab:crosslingual_embed}. Circularity \% from Table \ref{tab:experimental_results} is not included, as the error was too insignificant.

\subsection{Fine-Tuning Details}
\label{sec:fine_tuning_appendix}
We fine-tuned two models—Llama-3.1-8B and Llama-3.2-1B—on the dataset from Section \ref{sec:dataset} using parameter-efficient fine-tuning with LoRA \cite{Yu_2023}. Both models were trained with the same set of hyperparameters on an NVIDIA H100 GPU. For both models, we used LoRA with a rank of 64, LoRA alpha set to 16, and a dropout rate of 0.1. We enabled 4-bit quantization using the BitsAndBytes library \cite{dettmers2023qlora} with the nf4 quantization type and bfloat16 compute data type. The optimizer was paged\_adamw\_32bit with a learning rate of 2e-4, using an inverse square root learning rate schedule and a warmup ratio of 0.03. We used a batch size of 8 with gradient accumulation set to 1 and applied gradient clipping with a maximum norm of 0.3. The sequence length was limited to 512 tokens. 

\begin{table}
  \caption{Evaluation of NSM explications generated by LLMs for the benchmark set introduced in Section \ref{sec:dataset_evaluation_split}. We measure Explication Score (Equation \ref{eq:explicataion_score}, higher is better), Substitutability Score (Equation \ref{eq:average_substitutability_score}, higher is better), Legality Score (higher is better), Primes Ratio (higher is better), and Molecules Ratio (lower is better). Dictionary Definitions are existing definitions provided from WordNet. 
  Best underlined.
  While dictionary definitions and Llama-8B, achieve high substitutability, they fail to sufficiently use the NSM primes, lowering their legality and overall scores. DeepNSM models surpass SOTA general LLMs for NSM explication generation despite only having 1B and 8B parameters.}
  \label{tab:experimental_results_error_bars}
  \centering
  \begin{tabular}{lccccc
  }
  \toprule 
  \multicolumn{1}{c}{Model} & 
  \makecell{\textbf{Explication} \\ \textbf{Score} $\uparrow$}  & 
  \makecell{Sub. \\ Score $\uparrow$}  & 
  \makecell{Legality \\ Score $\uparrow$}  & 
  \makecell{Primes \\ Ratio $\uparrow$}  & 
  \makecell{Mols. \\ Ratio $\downarrow$}  
  \\
  \midrule
  Dictionary Def. & \textbf{13.4} $\pm$ .16 & \underline{12.14} $\pm$ .08 & -4.7 $\pm$ .0082 & 8.0 $\pm$ .0008 & 55.1 $\pm$ .0010 
  \\
  \midrule
  Llama-3.2-1B-it & \textbf{6.1} $\pm$ .09 & 6.82 $\pm$ .05 & -0.4 $\pm$ .01 & 29.2 $\pm$ .0008 & 33.2 $\pm$ .0008 
  \\
  Llama-3.1-8B-it & \textbf{20.0} $\pm$ .11 & 7.86 $\pm$ .05 & 2.3 $\pm$ .01 & 43.9 $\pm$ .0008 & 20.9 $\pm$ .0006 
  \\
  Gemini-2.0-Flash & \textbf{22.2} $\pm$ .10 & 6.40 $\pm$ .05 & 5.1 $\pm$ .01 & 62.1 $\pm$ .0006 & 11.5 $\pm$ .0004 
  \\
  GPT-4o & \textbf{22.9} $\pm$ .10 & 6.95 $\pm$ .05 & 4.6 $\pm$ .01 & 59.9 $\pm$ .0006 & 13.8 $\pm$ .0005 
  \\
  \midrule
  DeepNSM-1B$^\dagger$ & \textbf{19.7} $\pm$ .11 & 5.22 $\pm$ .05 & 5.0 $\pm$ .01 & 61.1 $\pm$ .0009 & 10.9 $\pm$ .0005 
  \\
  DeepNSM-8B$^\dagger$ & \textbf{22.2} $\pm$ .11 & 6.40 $\pm$ .05 & 4.9 $\pm$ .01 & 60.0 $\pm$ .0008 & 11.5 $\pm$ .0004 
  \\
  DeepNSM-1B & \textbf{23.2} $\pm$ .10 & 7.02 $\pm$ .05 & 5.1 $\pm$ .01 & 61.9 $\pm$ .0007 & 10.6 $\pm$ .0004 
  \\
  DeepNSM-8B & \textbf{\underline{24.6}} $\pm$ .10 & 7.34 $\pm$ .05 & \underline{5.4} $\pm$ .01 & \underline{63.9} $\pm$ .0007 & \underline{10.4} $\pm$ .0004 
  \\
  \bottomrule
\end{tabular}
  
  \vspace{0.5em}
  \raggedright
  {\footnotesize $^\dagger$Trained on a dataset with no quality filtering applied.}
\end{table}

\begin{table}
\caption{Cross-translatability BLEU scores ($\uparrow$) on model-generated explications for five low-resource languages from different families: Alur (alz), Kinyarwanda (rw), Dzongkha (dn), Dinka (din), and Abkhaz (ab). Best scores are underlined and bolded. }
\centering
\begin{tabular}{l|ccccc}
\toprule
& \multicolumn{5}{c}{BLEU $\uparrow$} \\
\multicolumn{1}{c|}{Model} & alz & rw & dn & din & ab \\
\midrule
Dictionary Def. & 22.9 $\pm$ .13 & 35.1 $\pm$ .16 & 19.6 $\pm$ .10 & 24.4 $\pm$ .13 & 26.9 $\pm$ .15 \\
\midrule
Llama-3.2-1B-it & 29.0 $\pm$ .10 & 45.0 $\pm$ .12 & 21.0 $\pm$ .07 & 25.8 $\pm$ .09 & 25.2 $\pm$ .08 \\
Llama-3.1-8B-it & 33.2 $\pm$ .10 & 50.8 $\pm$ .11 & \underline{23.8} $\pm$ .08 & 31.8 $\pm$ .09 & 27.8 $\pm$ .09 \\
Gemini-2.0-Flash & 33.5 $\pm$ .07 & 56.3 $\pm$ .10 & 23.2 $\pm$ .06 & 32.2 $\pm$ .07 & 30.1 $\pm$ .08 \\
GPT-4o & 31.2 $\pm$ .08 & 51.3 $\pm$ .11 & 21.6 $\pm$ .07 & 31.2 $\pm$ .09 & 28.7 $\pm$ .08 \\ 
\midrule
DeepNSM-1B & 34.5 $\pm$ .11 & \underline{56.9} $\pm$ .11 & 23.6 $\pm$ .07 & \underline{32.6} $\pm$ .09 & 31.9 $\pm$ .09 \\
DeepNSM-8B & \underline{37.0} $\pm$ .10 & 55.6 $\pm$ .10 & 23.3 $\pm$ .07 & 32.4 $\pm$ .08 & \underline{33.2} $\pm$ .09 \\
\bottomrule
\end{tabular}
\label{tab:crosslingual_bleu}
\end{table}

\begin{table}
\caption{Cross-translatability Embedding Similarity ($\uparrow$) on model-generated explications for five low-resource languages from different families: Alur (alz), Kinyarwanda (rw), Dzongkha (dn), Dinka (din), and Abkhaz (ab). Best scores are underlined and bolded.}
\centering
\begin{tabular}{l|ccccc}
\toprule
& \multicolumn{5}{c}{Embedding Similarity $\uparrow$} \\
\multicolumn{1}{c|}{Model} & alz & rw & dz & din & ab \\
\midrule
Dictionary Def. & 69.6 $\pm$ .13 & 79.6 $\pm$ .10 & 73.3 $\pm$ .11 & 69.1 $\pm$ .13 & 78.0 $\pm$ .10 \\
\midrule
Llama-3.2-1B-it & 79.3 $\pm$ .07 & 88.0 $\pm$ .05 & 81.3 $\pm$ .06 & 77.3 $\pm$ .08 & 84.2 $\pm$ .05 \\
Llama-3.1-8B-it & 81.6 $\pm$ .07 & 89.8 $\pm$ .06 & 81.8 $\pm$ .06 & 78.1 $\pm$ .07 & 84.0 $\pm$ .06 \\
Gemini-2.0-Flash & 84.0 $\pm$ .06 & \underline{93.0} $\pm$ .03 & 81.7 $\pm$ .06 & 78.5 $\pm$ .06 & 85.1 $\pm$ .05 \\
GPT-4o & 82.0 $\pm$ .05 & 91.4 $\pm$ .04 & 78.9 $\pm$ .07 & 77.4 $\pm$ .07 & 84.8 $\pm$ .05 \\
\midrule
DeepNSM-1B & 81.6 $\pm$ .07 & 91.4 $\pm$ .04 & 80.4 $\pm$ .06 & 78.0 $\pm$ .07 & 82.7 $\pm$ .07 \\
DeepNSM-8B & \underline{84.2} $\pm$ .05 & 91.7 $\pm$ .04 & \underline{82.0} $\pm$ .06 & \underline{79.0} $\pm$ .06 & \underline{85.4} $\pm$ .05 \\
\bottomrule
\end{tabular}
\label{tab:crosslingual_embed}
\end{table}

\subsection{Example Model Outputs}
\label{sec:example_model_outputs_appendix}
In this section, we present representative outputs from the models evaluated in Section \ref{sec:model_experiments}. For each example, we include the target word, the accompanying example sentences, and the corresponding explications generated by each model. We then conduct a qualitative analysis of these explications, highlighting in green the one we consider the most effective.

\begin{figure}[h!]
    \centering
    \includegraphics[width=1\linewidth]{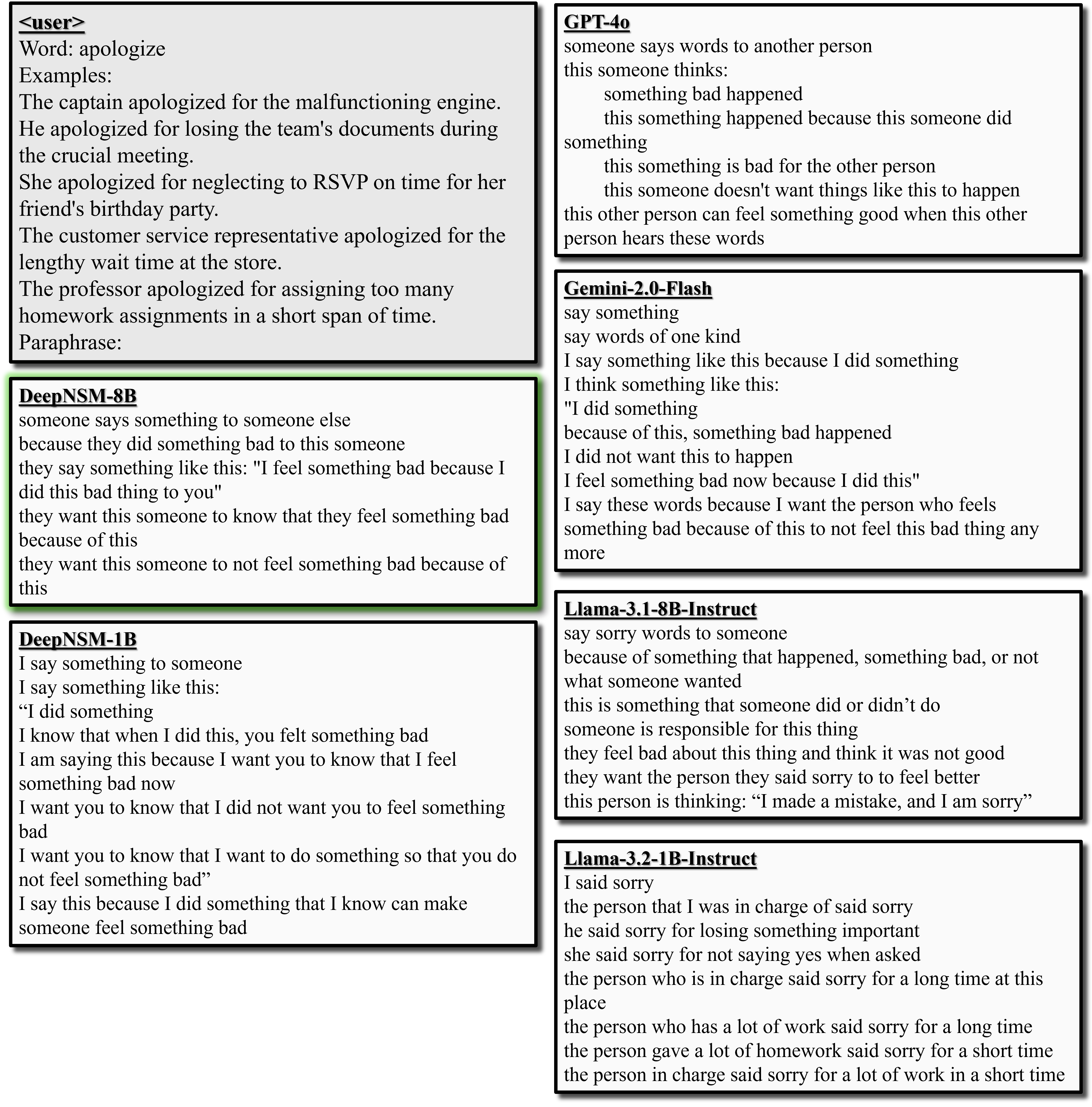}
    \caption{
    Model-generated explications for the word "apologize." 
    Both the Llama-3.1-8B-Instruct and Llama-3.2-1B-Instruct models clearly fail to utilize the semantic primes, introducing unnecessary words like "responsible," "homework," "mistake," and containing circularity ("sorry words"). 
    GPT-4o's explication successfully identifies that something bad happened and that the speaker wants the listener to feel better. However, it fails to clearly express the speaker’s regret—something like “I feel bad because I did something bad to you”—which is essential for an apology. It also omits that the speaker is saying these words specifically to communicate this feeling of regret (make the other someone "know" they "feel bad").
    The explication from Gemini-2.0 Flash is slightly redundant in its opening lines and similarly falls short in expressing the speaker’s internal state. Like GPT-4o, it focuses on wanting the listener to stop feeling bad, rather than emphasizing the speaker’s desire to communicate their own regret.
    DeepNSM-1B captures most of the core components of an apology. It reflects a past action that harmed the listener ("I did something ... you felt something bad"), the speaker’s awareness of this harm, and their emotional response of regret ("I feel something bad because I did this bad thing to you"). It also notes the speaker’s desire to communicate this feeling ("I want you to know that I feel something bad now"). However, it introduces an extra element about doing (not just saying) something to change the listener’s feelings, which slightly weakens the clarity of the apology by implying a broader corrective intent.
    DeepNSM-8B builds on the strengths of 1B while avoiding this ambiguity. It clearly conveys the past harmful action, the speaker’s acknowledgment and regret, and the intent to communicate this regret explicitly in order to make the hearer feel better. We believe DeepNSM-8B provides the most accurate and complete explication for this word.
    }
    \label{fig:model_outputs_apologize}
\end{figure}

\begin{figure}[h!]
    \centering
    \includegraphics[width=1\linewidth]{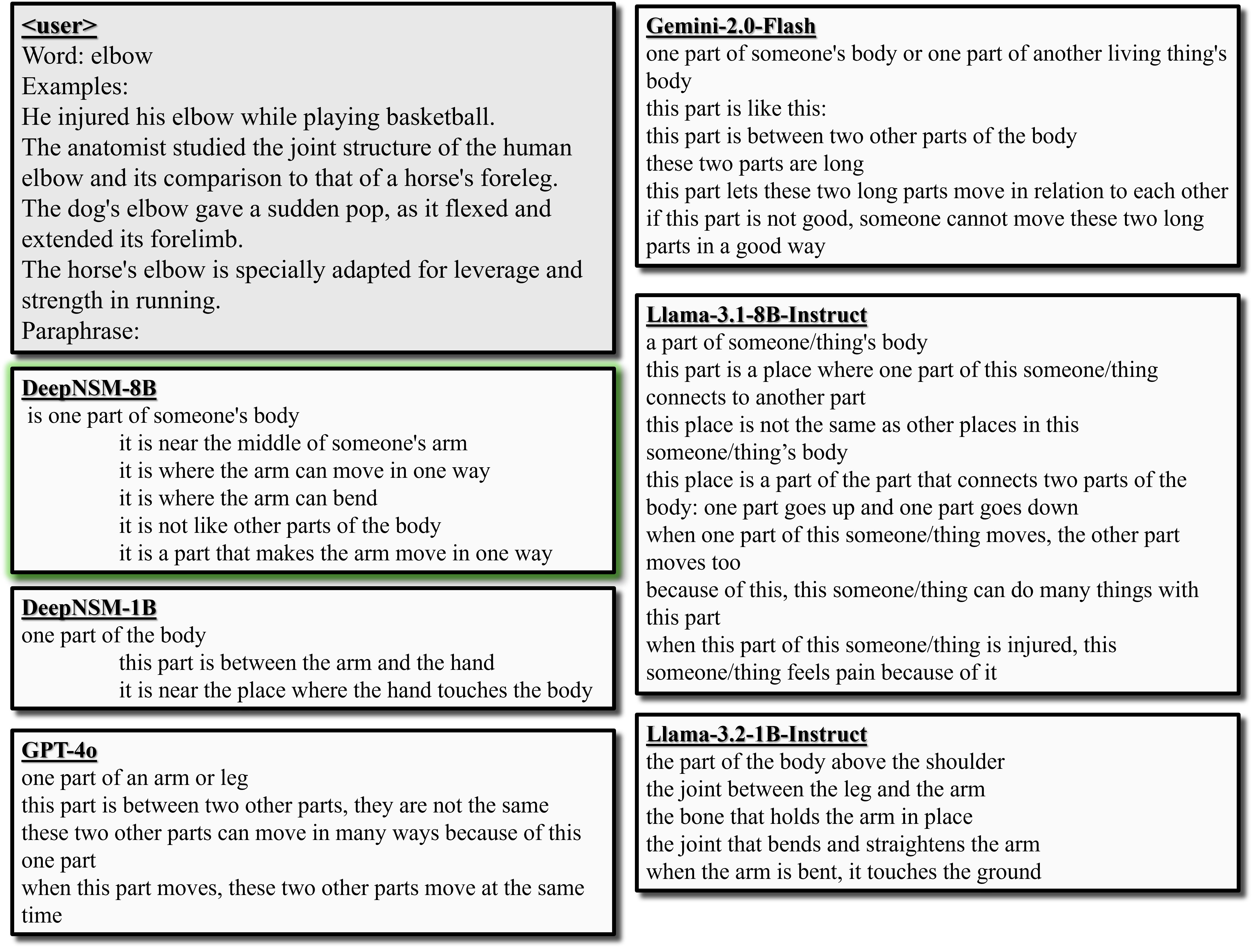}
    \caption{
    Model-generated explications for the word "elbow."
    Similarly to Figure \ref{fig:model_outputs_apologize}, the Llama models again produce noticeably lower-quality outputs than the rest of the group. While Llama-3.1-8B makes slightly better use of the primes, its description is vague and includes imprecise and potentially misleading information, for example, stating that “one part goes up and one part goes down,” which doesn’t clearly characterize what an elbow does.
    GPT-4o’s explication has several issues. It introduces the term “leg,” likely influenced by the animal examples (e.g., dogs and horses), but an NSM explication should typically define “elbow” specifically in terms of its relation to the human body. The phrase “when this part moves, these two other parts move at the same time” is also vague and fails to clearly capture what the elbow does.
    Gemini-2.0 Flash’s description of the elbow as being “between two long parts” is too vague, and likely is closer to an explication for "knee," making the explication ambiguous and not clearly specific to the concept of “elbow.”
    While DeepNSM-1B provides a much more concise description, locating the elbow at “where the hand touches the body,” is not accurate.
    We identify DeepNSM-8B as providing the clearest and most accurate explication for this word. It introduces simple molecules like “arm,” “middle,” and “bend” that are comparable in complexity to those used by other models, but it uses them more effectively to clearly convey both the location of the elbow and what it does.
    }
    \label{fig:model_outputs_elbow}
\end{figure}

\begin{figure}[h!]
    \centering
    \includegraphics[width=1\linewidth]{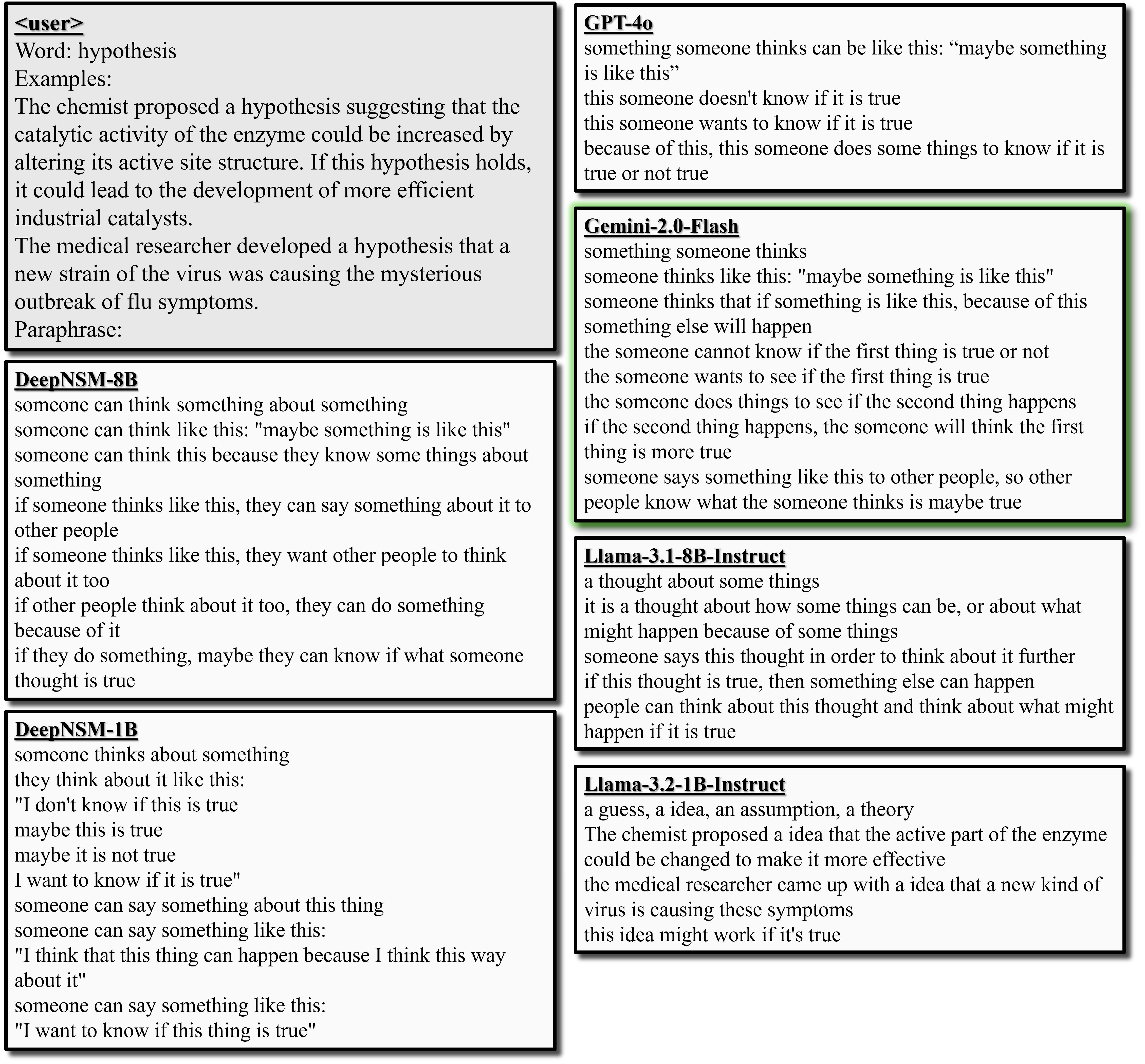}
    \caption{
    Model-generated explications for the word "hypothesis." 
    The word hypothesis refers to a proposition about something unknown that can be tested or disproven through evidence. As in earlier examples, the Llama model outputs can be dismissed. GPT-4o’s explication fails to clearly convey falsifiability based on evidence—its phrase “doing some things to know if it is true or not true” lacks the idea that this determination depends on whether certain things happen or not. DeepNSM-8B makes the same omission. DeepNSM-1B’s output is weaker still: it does not clearly present the initial proposition (e.g., “this thing is like this”) and fails to include any sense of testing or “doing” to evaluate it. We identify Gemini 2.0 Flash as offering the strongest explication for this word. It captures the predictive structure of a hypothesis (“if something is like this, something else will happen”), the testing process (“someone does things to see”), and the dependence on evidence for evaluation (“if this happens, the someone will think the first thing is more true”).}
    \label{fig:model_outputs_hypothesis}
\end{figure}

\begin{figure}[h!]
    \centering
    \includegraphics[width=1\linewidth]{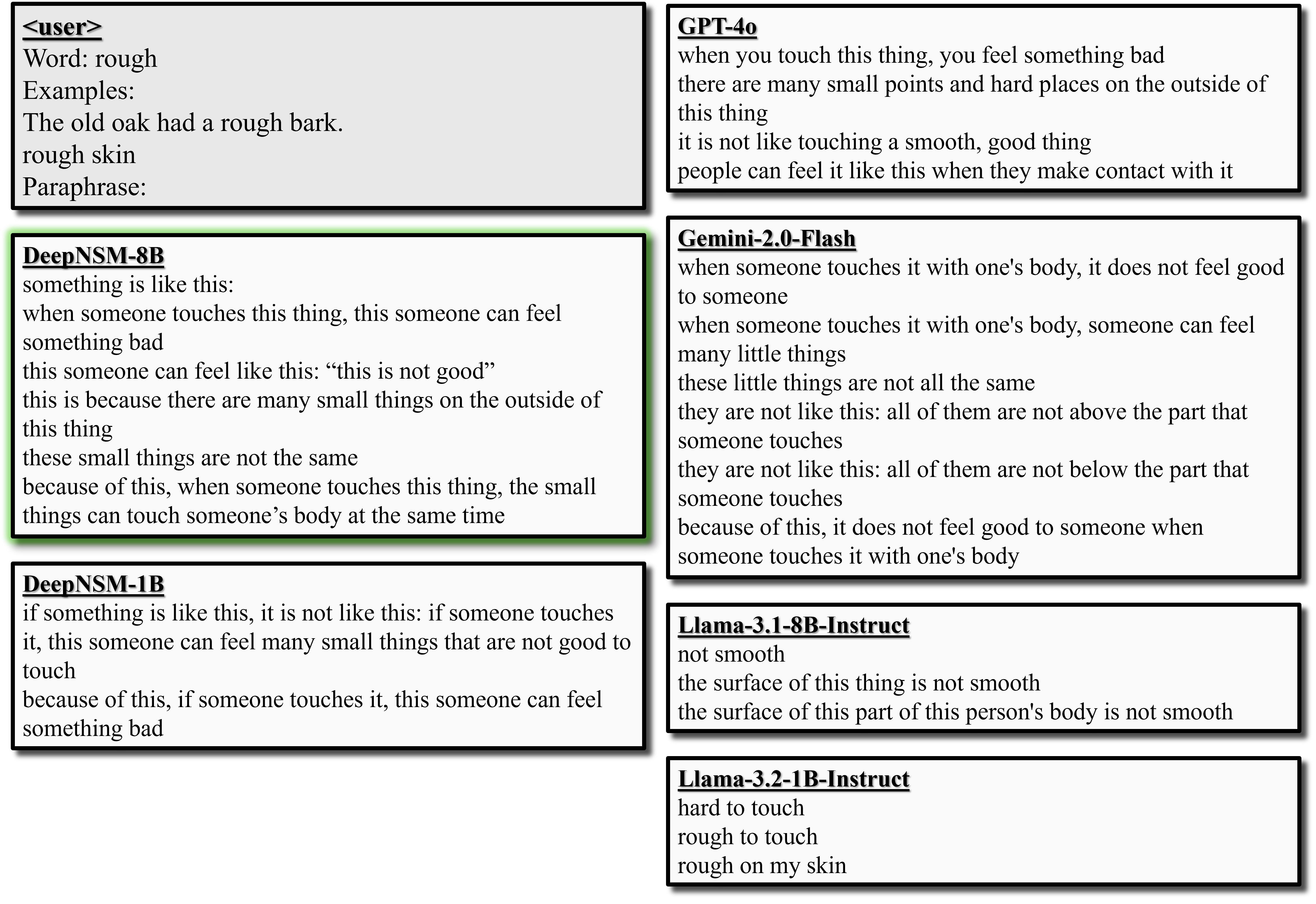}
    \caption{
    Model-generated explications for the word "rough." 
    As with previous examples, the Llama-generated explications can be clearly set aside. GPT-4o’s version introduces non-prime terms like “contact” and “points,” which add unnecessary complexity, but it still successfully conveys the unpleasant feeling caused by touching a surface with many small, hard projections. Gemini-2.0 Flash attempts to express a similar idea, but its phrasing—such as referencing things “not above/below the part”—introduces confusion, as these spatial qualifiers don’t clearly relate to the sensation of roughness. DeepNSM-1B includes a critical error by stating that things like this do not have small things on the outside, which is the opposite of what “rough” means. This contradiction makes the explication ineffective. In contrast, DeepNSM-8B avoids these issues: it captures the same descriptive meaning as GPT-4o without introducing non-prime terms or misleading spatial language. We identify DeepNSM-8B’s explication as the most accurate and effective for this word.
    }
    \label{fig:model_outputs_rough}
\end{figure}

\clearpage

\section{Prompts}
\label{sec:prompts_appendix}
\begin{figure}[h!]
    \centering
    \includegraphics[width=1\linewidth]{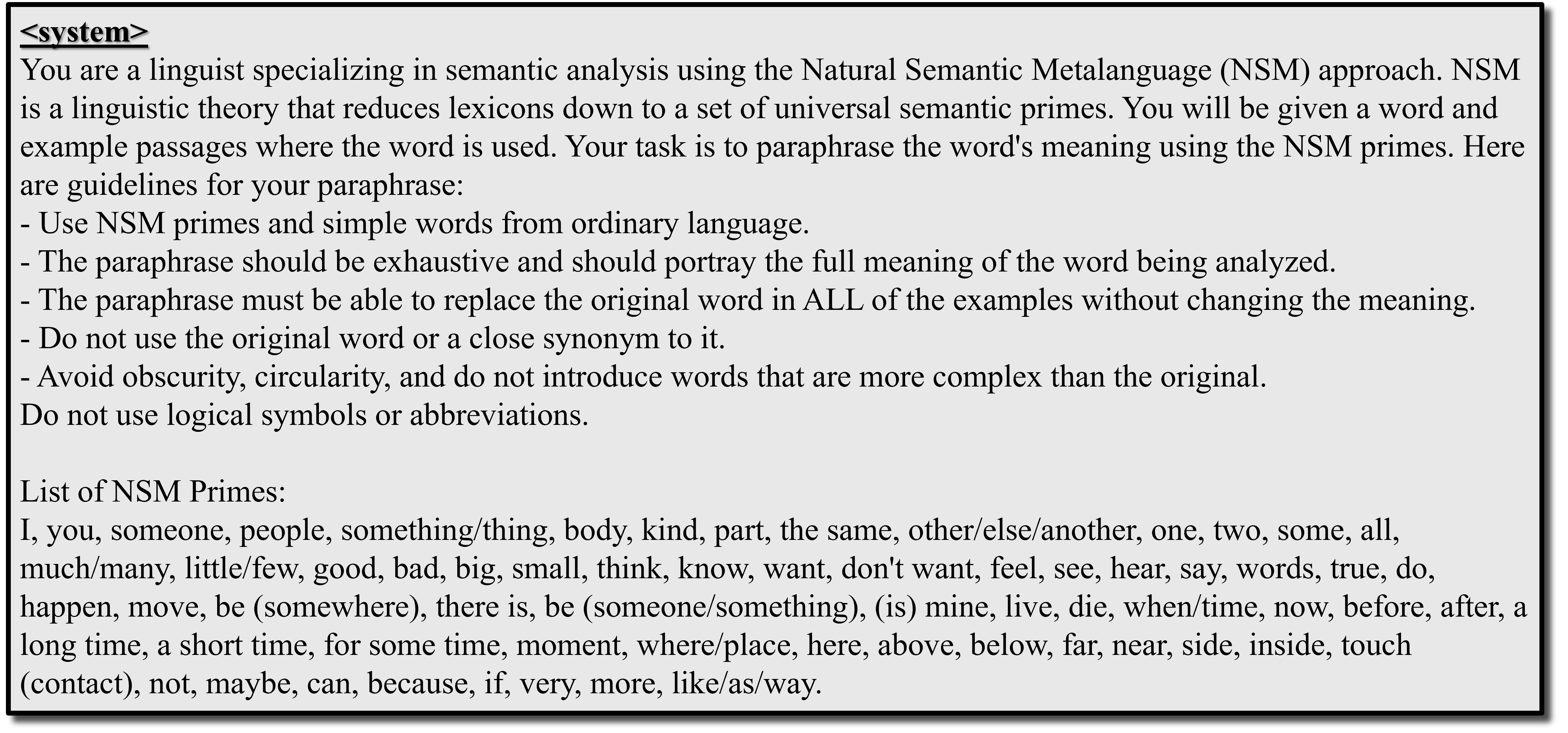}
    \caption{Explication Generation System Prompt.}
    \label{fig:explication_generation_sys}
\end{figure}
\begin{figure}[h!]
    \centering
    \includegraphics[width=1\linewidth]{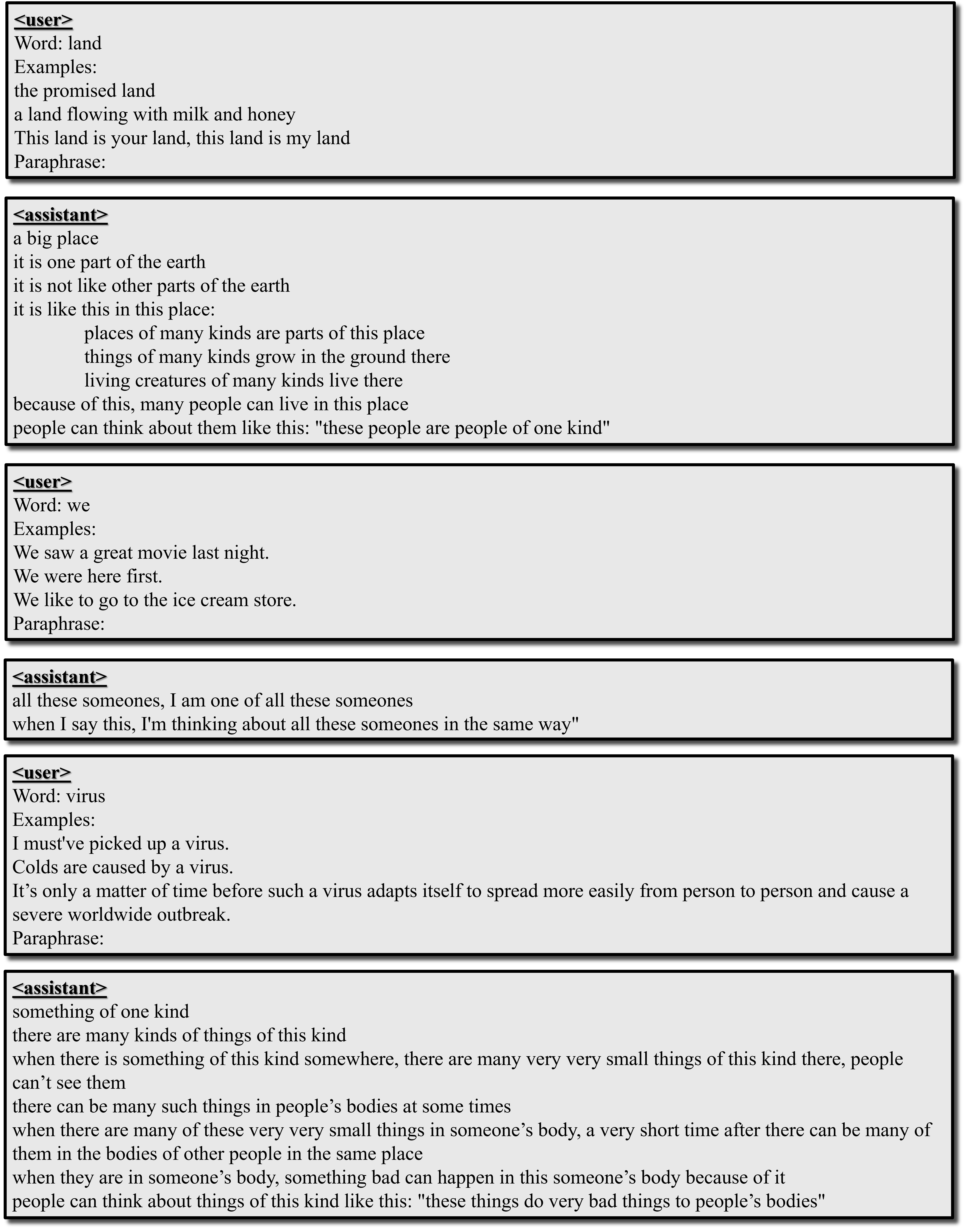}
    \caption{Few-Shot Examples used for the Explication Generation Prompt.}
    \label{fig:explication_generation_examples}
\end{figure}


\clearpage

\section{Extending This Work to Other Languages}
\label{sec:extending_work}
While this work focuses on paraphrasing English texts into the semantic primes, the methods we propose are readily extendable to other high-resource languages. Even in cases where parallel corpora are limited, the approach remains viable as long as the underlying language models have been trained on sufficient text in the target language. This opens up immediate opportunities for researchers and practitioners aiming to translate text in a high-resource language into a low-resource one. Specifically, high-resource paraphrases into semantic primes can serve as an intermediate representation that is more easily translated into low-resource languages, potentially reducing ambiguity and improving semantic fidelity.

Additionally, our work opens valuable opportunities for paraphrasing texts written in low-resource languages (LRLs) into the semantic primes—a task that is very challenging due to limited data and linguistic documentation. One promising direction is to leverage English-to-primes models like DeepNSM to first convert large English corpora into semantic primes. These prime-level representations can then be used to guide models training on LRLs, especially once the primes have been identified and validated in the target language. For example, models could learn to paraphrase unknown words in the LRL by analyzing how they appear near known prime expressions—effectively applying an informed form of distributional semantics grounded in semantic primes. However, this approach risks introducing biases from English-specific prime usage patterns, highlighting the need for further cross-linguistic study.

Alternatively, the evaluation methods proposed in this work can serve as optimization signals or auxiliary objectives for training or fine-tuning models directly on LRL text. In cases where training data is sparse, fluent speakers of the target language could also use the paraphrased dataset as a foundation, combining our automatic techniques with human annotation to refine and validate prime-level paraphrases, leading to high-quality NSM datasets in a wide range of languages.

\end{document}